\documentclass[10pt,journal,compsoc]{IEEEtran}
\usepackage[nocompress]{cite}

\usepackage{mathrsfs}
\usepackage{amsfonts}
\usepackage{graphicx,epsfig,amssymb,amsmath}
\usepackage{color,xcolor}
\definecolor{red}{rgb}{1.00, 0.00, 0.00}  
\usepackage{pifont}
\usepackage{stmaryrd}
\usepackage{setspace}
\usepackage{subfigure}
\usepackage{array}
\usepackage{float}
\usepackage{multirow}
\usepackage{algorithmic,algorithm}
\usepackage{epstopdf}
\usepackage{mathtools}
\usepackage{diagbox}
\usepackage{booktabs}
\usepackage{verbatim}

\newcommand{\bm}[1]{\mbox{\boldmath{$#1$}}}
\newcommand{\tabincell}[2]{\begin{tabular}{@{}#1@{}}#2\end{tabular}}

\usepackage{arydshln}
\usepackage{enumitem}
\usepackage{ragged2e}
\usepackage{hyperref}
\begin{document}
	
\title{A Fast Graph Neural Network-Based Method for Winner Determination in Multi-Unit Combinatorial Auctions}

\author{Mengyuan~Lee,~ Seyyedali~Hosseinalipour,~\IEEEmembership{Member,~IEEE,} Christopher~G.~Brinton,~\IEEEmembership{Senior Member,~IEEE,} Guanding~Yu,~\IEEEmembership{Senior Member,~IEEE,} and~Huaiyu~Dai,~\IEEEmembership{Fellow,~IEEE}
	\IEEEcompsocitemizethanks{
		\IEEEcompsocthanksitem M. Lee  and  G. Yu are with College of Information Science and Electronic Engineering, Zhejiang University, Hangzhou 310027, China. \protect\\
		E-mail: \{mengyuan\_lee,  yuguanding\}@zju.edu.cn.
		\IEEEcompsocthanksitem S. Hosseinalipour and C. Brinton are with the School of Electrical and Computer Engineering, Purdue University,
		West Lafayette, IN 47907, USA. \protect\\
		E-mail: \{hosseina,cgb\}@purdue.edu.
		\IEEEcompsocthanksitem H. Dai is with the Department of Electrical and Computer Engineering, North Carolina State University,
		Raleigh, NC 27606, USA.\protect\\
		E-mail: hdai@ncsu.edu
		\IEEEcompsocthanksitem This work was done while M. Lee was a visiting PhD student at NC State University.
}}

\IEEEtitleabstractindextext{
\begin{abstract}
\justifying  
The combinatorial auction (CA) is an efficient mechanism for resource allocation in different fields, including cloud computing. It can obtain high economic efficiency and user flexibility by allowing bidders to submit bids for combinations of different items instead of only for individual items. However, the problem of allocating items among the bidders to maximize the auctioneers' revenue, i.e., the winner determination problem (WDP), is NP-complete to solve and inapproximable. Existing works for WDPs are generally based on mathematical optimization techniques and most of them focus on the single-unit WDP, where each item only has one unit. On the contrary, few works consider the multi-unit WDP in which each item may have multiple units. Given that the multi-unit WDP  is more complicated but prevalent in cloud computing, we propose leveraging machine learning (ML) techniques to develop a novel low-complexity algorithm for solving this problem with negligible revenue loss. Specifically, we model the multi-unit WDP as an augmented bipartite bid-item graph and use a graph neural network (GNN) with half-convolution operations to learn the probability of each bid belonging to the optimal allocation. To improve the sample generation efficiency and decrease the number of needed labeled instances, we propose two different sample generation processes. We also develop two novel graph-based post-processing algorithms to transform the outputs of the GNN into feasible solutions. Through simulations on both synthetic instances and a specific virtual machine (VM) allocation problem in a cloud computing platform, we validate that our proposed method can approach optimal performance with low complexity and has good generalization ability in terms of problem size and user-type distribution.
\end{abstract}

}

\maketitle

\IEEEraisesectionheading{\section{Introduction}}
\IEEEPARstart{A}{n} auction involves allocating items or resources to a group of bidders. Given complementarity and substitution relations among available items, the combinatorial auction (CA) has been proposed to allow the bidders to submit bids for combinations of different items instead of only for individual items, which enhances  economic efficiency and user flexibility \cite{CA_survey}. CA has been used as an effective method for \textit{resource allocation} problems in a variety of fields, such as transportation services  \cite{transportation}, airport slot allocations  \cite{airport}, cognitive radio networks \cite{cognitive1,cognitive2}, grid systems \cite{grid}, mobile edge computing  \cite{mec}, and fog computing \cite{fog}. Moreover, many existing works in cloud computing have turned to the CA mechanism to solve  resource allocation problems for cloud computing platforms \cite{cloud1,cloud2,cloud3,cloud4}.

Generally, a CA consists of three steps: bidding, winner determination, and payment computation. First, each bidder submits its bid to the auctioneer, where each bid comprises the number of units that the bidder requests for each item and the maximum price that the bidder is willing to pay for the bundle. Then, the auctioneer allocates items among bidders aiming to maximize its own revenue. Finally, the auctioneer computes the payments of each bidder who would receive its requested bundle of items to ensure  certain properties for the auction mechanism, e.g., truthfulness \cite{CA_survey}.  Among the aforementioned steps, the most difficult and time-consuming step is the second one, which is usually called the \emph{winner determination problem} (WDP). The WDP belongs to the family of NP-complete problems and is inapproximable \cite{np_complete}. Existing algorithms for the WDP can be categorized into three types: (i) those that try to accelerate the process of finding the optimal allocation \cite{CASS,data_distribution,BOB,CABOB}, (ii) those that aim to develop low-complexity heuristic algorithms \cite{SS,Casanova}, and (iii) those that focus on solving some special cases \cite{m3_1,m3_2}. Moreover, most of the existing algorithms are developed for the single-unit CA, where each item only has one copy or unit.

In this paper, we are interested in the multi-unit CA, where each item has multiple units. It is prevalent in cloud computing. However, some algorithms mentioned above \cite{CASS,BOB,CABOB,m3_1}  for the single-unit CA make use of the property that there is only one unit for each item. Therefore, they cannot be easily generalized to WDPs in the multi-unit CA. Motivated by this, we will leverage machine learning (ML) techniques to develop efficient algorithms to solve the WDPs in the multi-unit CA. For simplicity, in the rest of this paper, the WDPs in single-unit and multi-unit CAs are referred to as single-unit and multi-unit WDPs, respectively.

\subsection{Related Work}
We are inspired by the success of ML in solving different NP-hard problems, e.g.,  \cite{hehe,minlp,bi_gnn,dl_bb,gbd,guided_tree,TSP,graphnn,wcl}. In these works, ML is generally employed for two purposes. On the one hand, it has been used to replace heavy computations in existing algorithms by fast approximations. For example, the authors in \cite{hehe,minlp,bi_gnn,dl_bb} have made use of different ML techniques, such as imitation learning, deep neural networks (DNNs), and graph neural networks (GNNs), to replace the pruning or branching strategies in the branch-and-bound algorithm, a widely-used algorithm with exponential complexity \cite{bb}. On the other hand, ML has been used to develop new low-complexity heuristic algorithms for NP-hard problems. For example,  the authors in \cite{guided_tree,TSP,graphnn,wcl} have proposed ML-based heuristic algorithms for the maximum independent set (MIS) problem, the traveling salesperson problem (TSP), the wireless link scheduling problem, and the linear sum assignment problem, respectively. Specifically, the authors have  trained the ML models to learn a mapping to the good solution by using the optimal/sub-optimal solutions of the corresponding NP-hard problems as training labels. Recently,  these two types of ML approaches have also been widely used in cloud computing  to solve  NP-hard problems, such as allocating CPU or memory resources \cite{cloud_ml2}, green scheduling algorithms for cloud servers \cite{cloud_ml3}, and joint virtual machine (VM) allocation and power management \cite{cloud_ml4}.

\subsection{Challenges and Overview of Methodology}
To solve the multi-unit WDP with ML techniques, we must overcome several challenges. First, we must develop a methodology that can encode and learn over the general WDP efficiently. To do this, we first propose an augmented bipartite bid-item graph for the multi-unit WDP with appropriate feature design. Then, we leverage GNNs that can capture the information in non-Euclidean structures  \cite{gnn_survey1,gnn_survey2} to learn the probability of each bid belonging to the optimal allocation.  In particular, our GNN structure employs half-convolution operations, designed for bipartite graphs to mitigate the complexity of the training process.  Next, we must generate labeled samples for the GNN to learn over, which are difficult to acquire at scale. Motivated by this, we propose two new methods to enhance the sample generation efficiency and decrease the number of needed labeled instances. Finally, we must transform the outputs of GNN into feasible solutions of the multi-unit WDP. To this end,  we develop two novel graph-based post-processing algorithms, which achieve different trade-offs between the time complexity and the revenue loss. 

We conduct extensive simulations on both synthetic instances and the specific VM allocation problem in cloud computing to verify the effectiveness and generalization ability of the proposed method. The results reveal that our proposed method can approach optimal performance with low complexity. Moreover, our method outperforms the state-of-the-art optimization software CPLEX\footnote{https://www.ibm.com/analytics/cplex-optimizer} in terms of time complexity while only experiencing a small revenue loss, and also outperforms some widely-used heuristic algorithms in terms of time complexity, revenue loss, resource utilization, and user satisfaction. Furthermore, the proposed method demonstrates good generalization ability in terms of problem size and distribution for different types of cloud computing users. As a result, our method can be trained rapidly with small-size instances that are easy to solve, and then be used to solve other similar instances with larger sizes and different user-type distributions, which is a preferred property in practice.

\subsection{Contributions and Outline}
To summarize, our main contributions are as follows.
\begin{itemize}[leftmargin=3mm]
	\item  We propose an augmented bipartite bid-item graph to model the multi-unit WDP. Based on this graph structure, we utilize a GNN model with half-convolution operations to mitigate the training complexity.
	\item We propose two novel sample generation methods for GNN model training, the optimum-only  and the mixed sample generation processes, which enhance the sample generation efficiency and the generalization ability of the model in terms of problem size.
	\item We develop two novel graph-based post-processing algorithms, the basic and the traversal post-processing algorithms, to interpret the GNN outputs in a manner suitable for time- and revenue-sensitive scenarios, respectively.
	\item We conduct extensive simulations through which we reveal that our method outperforms the  state-of-the-art optimization solver and has good generalization ability.
\end{itemize}

\vspace{2mm}
The rest of this paper is organized as follows. In Section \ref{s2}, we formalize the multi-unit WDP. In Section  \ref{s3}, we develop our GNN-based method for it, which includes the graph representation model, the GNN structure, the sample generation processes, and the graph-based post-processing algorithms. Extensive testing results on the synthetic instances and the VM allocation problem are presented in Sections  \ref{s4} and  \ref{s5}, respectively. Finally, we conclude the paper and provide directions for future work in Section \ref{s6}.

\section{Multi-Unit Winner Determination Problem} \label{s2}
The WDP aims to maximize the revenue of the auctioneer by allocating items to a subset of bidders. In this paper, we consider a general scenario where the auctioneer offers a set of $N$ items, $\{\iota_1, \iota_2, ..., \iota_N\}$, to sell. For item $\iota_n$, the auctioneer possesses $u_n$ available units. Meanwhile, there are a set of $M$ bidders denoted by $\{b_1, b_2, ..., b_M\}$, which are single-minded, i.e.,  each bidder only submits one bid.\footnote{Note that although the assumption of single-minded bidders is imposed,  the bidder can express requests on different bundles of items through a sequence of auctions.}   Each bid, $B_m$, is a tuple, $ \{(\lambda_m^1, \lambda_m^2, ..., \lambda_m^N), p_m\}$, where $\lambda_m^n$ is the number of units of item $\iota_n$ that bidder $b_m$ requests, and $p_m>0$ is the bidder's valuation for this bundle, i.e., the maximum price that the bidder is willing to pay for it.

We introduce a binary allocation vector, $\bm{a}=[a_1, a_2, ..., a_M]$, to indicate the results of the WDP, where $a_m=1$ indicates that the requested bundle of bidder $b_m$ is satisfied and $a_m=0$ implies the opposite. Under this model, the requested bundle of each bidder cannot be partially satisfied. The WDP is then formulated  as the following optimization problem:
\begin{eqnarray*}\label{WDP}
	\hspace{7.2em}\underset{\bm{a}}{\textrm{maximize}}~ \sum_{m=1}^M p_m a_m,\hspace{8em}
	\eqref{WDP_obj} \label{WDP_obj}
\end{eqnarray*}
\begin{subequations}
	subject to
	\begin{align}
	\sum_{m=1}^M \lambda_m^n a_m \leq u_n, n=1, 2, ..., N,\label{WDP_sub1}
	\end{align}
	\vspace{-1em}
	\begin{align}
	a_{m} \in \{0,1\}, m=1, 2, ..., M,\label{WDP_sub2}
	\end{align}
\end{subequations}
where the objective function is the auctioneer's revenue, and constraint (\ref{WDP_sub1}) guarantees that the sum of allocated units of any item $\iota_n$ over all the winning bids does not exceed its available units $u_n$. If there is only one unit for each item, i.e., $u_1=u_2=...=u_N=1$, the auction is called single-unit CA and Problem (\ref{WDP}) reduces to single-unit WDP, which is equivalent to the weighted set packing problem \cite{mit_book}. Otherwise, the auction is called multi-unit CA and Problem (\ref{WDP})  is dubbed multi-unit WDP. 

In general, Problem (\ref{WDP})  is difficult to solve due to its NP-completeness \cite{np_complete}. Existing works tackle this problem based on one of the following three approaches:
\begin{itemize}[leftmargin=3mm]
    \item developing techniques to decrease the running time of finding the optimal allocation, such as combinatorial auction structured search\cite{CASS}, combinatorial auction multi-unit search \cite{data_distribution}, branch on bids (BoB) \cite{BOB}, and combinatorial auction BoB \cite{CABOB};
    \item developing heuristic algorithms to efficiently find sub-optimal solutions, such as  relaxed linear programming based method (RLP) \cite{cloud1,SS}, shadow surplus (SS) \cite{SS}, and Casanova \cite{Casanova};
    \item investigating special cases where optimal solutions can be obtained efficiently \cite{m3_1,m3_2}.
\end{itemize}
Most of the aforementioned works focus on the single-unit WDP \cite{CASS,BOB,CABOB,SS,Casanova,m3_1}. The algorithms for the weighted set packing problem that is well studied can be also applied to the single-unit WDP. However, nearly all algorithms for the single-unit WDP make use of the property that each item only has one unit, and thus cannot be generalized to the multi-unit WDP. On the contrary, few works focus on the multi-unit WDP \cite{data_distribution,cloud1,m3_2}, which is more challenging to solve and difficult to transform into problems with well studied solutions. 

Existing works for the multi-unit WDP have employed different mathematical optimization techniques. For example, the authors in \cite{cloud1} have proposed to relax the binary allocation vector to solve the multi-unit WDP, while the authors in \cite{data_distribution} have accelerated the branch-and-bound searching process by tailoring the upper bound function specifically to the multi-unit WDP. Given the complexity of the multi-unit WDP and its prevalence in the cloud computing domain \cite{cloud1,cloud2,cloud3,cloud4}, in this paper, we propose solving it via a novel GNN-based methodology. Different from theses existing methods, our proposed method is data driven and does not rely on mathematical optimization techniques. As we will show in Section \ref{s4_4}, the proposed data-driven method outperforms the existing works in terms of time complexity, revenue loss, resource utilization, and user satisfaction.

\section{GNN-Based Method for Multi-Unit WDP} \label{s3}
In this section, we develop the GNN-based method for Problem (\ref{WDP}). We first provide an overview of the proposed method. Then we design the graph representation and develop the GNN structure. Next, we discuss the sample generation and training processes, and develop two graph-based post-processing algorithms that transform the outputs of GNN into feasible solutions of Problem (\ref{WDP}). Finally, we analyze the time complexity of the proposed method.

\subsection{Overview of the GNN-Based Approach} \label{s3_1}
As the name implies, GNNs are neural networks for graphs defined over the non-Euclidean space. They are closely related but different from the popular convolutional neural networks (CNNs). CNNs have a much smaller number of parameters than traditional fully-connected DNNs since they replace full linear operations at each layer with shared convolutional filters. These convolutional filters can only operate on the grid-like data over the regular Euclidean space. To find a generalization of CNNs to deal with graphs over the non-Euclidean space, GNNs are proposed to aggregate information from graph structures \cite{gnn_survey1,gnn_survey2}. Specifically, for a graph $G(\mathcal{V},\mathcal{E})$ with the node set $\mathcal{V}$ and the edge set $\mathcal{E}$, GNNs aim at learning a state embedding $\bm{h}_v  \in \mathbb{R}^q$ of dimension $q$ for each node  $v \in \mathcal{V}$ by using the graph topology information, node features, and edge features. Note that $q$ is a hyperparameter whose value is manually set. Then the state embedding $\bm{h}_v$ is used to produce the output $\bm{o}_v$ of node $v$. 

Traditional DNNs generally consist of layers of neurons. In contrast, GNNs involve successive passes of graph convolution operations, which take node features, edge features, and the graph adjacency matrix as inputs. For each pass, there is a local transition function and a local output function instead of the convolutional filters in CNNs. The local transition function is shared among all nodes in $\mathcal{V}$ and collects neighborhood information to update the state of each node. On the other hand, the local output function, also shared by all nodes, defines how the output is produced for each node. Specifically, for the $k$-th pass, the hidden state of node $v$, denoted as $\bm{h}_v^{(k)}$, is first updated by the local transition function, $f_t^{(k)}$, as
\begin{equation}
\bm{h}_v^{(k)} = f_t^{(k)}(\bm{x}_v,\bm{x}_{co[v]},\bm{x}_{ne[v]}, \bm{h}^{(k-1)}_{ne[v]}), \label{local_tran}
\end{equation}
where $\bm{h}^{(k-1)}_{ne[v]}$ are the hidden states of the neighbors of node $v$ at the $(k-1)$-th pass. Also,
$\bm{x}_v$, $\bm{x}_{co[v]}$, $\bm{x}_{ne[v]}$ are the features of node $v$, its edges, and its neighbors, respectively. Then the output of node $v$ at the $k$-th pass, denoted as $\bm{o}_v^{(k)}$, is computed by using the local output function, $f_o^{(k)}$, as
\begin{equation}
\bm{o}_v^{(k)} = f_o^{(k)}(\bm{x}_v,\bm{h}^{(k)}_{v}).\label{local_out}
\end{equation}
Generally, the local transition and output functions at each pass are realized through fully-connected neural networks. More details will be given in Section \ref{s3_3}.

Moreover, GNNs have some special properties which make them a good choice for the multi-unit WDP. First, they are well-defined regardless of the input size, which implies that a trained GNN model can be used for graphs with different sizes. As for the multi-unit WDP, we can train models from small-size WDPs and use it to solve large-size WDPs. In this way, we avoid the training process for large-size problems, which is both time- and memory-consuming. Second, they are permutation-invariant, implying that the output of GNN is not influenced by the order in which the nodes are presented. In our cases, the allocation will remain the same regardless of the indices or the order of the bids.

In the following, we propose an efficient heuristic algorithm for the multi-unit WDP by using the aforementioned GNN. Specifically, we first propose a graph model for  Problem (\ref{WDP}) and then use GNNs to obtain an output for each bid. Given that the allocation vector, $\bm{a}$, is binary, we set the output of the GNNs  to be a continuous probability map in $[0,1]^M$ that indicates the probability of each bid belonging to the optimal allocation. After getting the predicted probability map, we develop two graph-based post-processing algorithms to transform the probability map into a feasible binary allocation result of the multi-unit WDP.  In the remaining parts of this section, we will introduce more details for each of the steps mentioned above.

\subsection{Graph Representation for Multi-Unit WDP}\label{s3_2}
In literature, there are several existing graph representation methods for WDPs \cite{mit_book}. The first one is the bid graph, where each node represents a bid and two nodes are adjacent if and only if they have an item in common. The second one is the item graph, where each node represents an item. Moreover, an item graph is valid if, for every bid, the items contained in that bid constitute a connected graph. Generally, there exist different valid item graphs for a specific  WDP. The third one is the bid-item graph, which is a bipartite graph denoted as $G(\mathcal{V}_b, \mathcal{V}_\iota, \mathcal{E})$. Specifically, each node in $\mathcal{V}_b$ represents a bid, and each node in $\mathcal{V}_\iota$ represents an item. There is an edge, $e(b_m,\iota_n) \in \mathcal{E}$, between the bid node $v_{b_m}$ and the item node $v_{\iota_n}$ if item $\iota_n$ is requested by bidder $b_m$. 

Among the aforementioned three methods, the first two are suitable for the single-unit WDP. They mainly reflect the relations between different bids/items. However, for a multi-unit WDP, the number of units that each bid requests for each item, $\lambda_m^n$, is indispensable. Both bid graph and item graph fail to include this information. On the contrary, $\lambda_m^n$ can be easily included in the bid-item graph as the edge weights. Therefore, we use the bid-item graph to represent the original multi-unit WDP in (\ref{WDP}).

We further design features for the nodes and the edges to incorporate all needed information for solving a multi-unit WDP into the graph. As mentioned before, the outputs of the GNNs are determined by both the topology information and the features of nodes and edges. Therefore, proper feature design is crucial for the final performance. 

\begin{figure}
	\centering
	\includegraphics[width=0.9\linewidth, height=0.18\textheight]{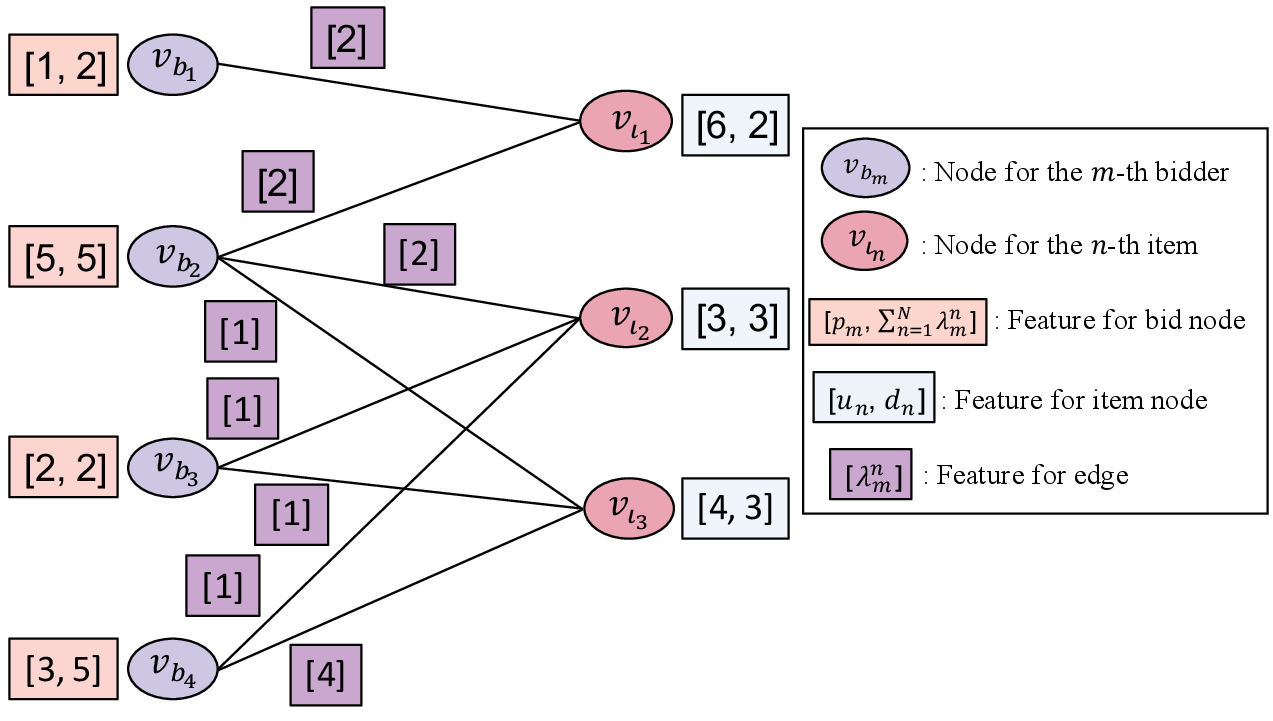}
	\caption{The augmented bid-item graph and designed features for a specific multi-unit WDP instance with $M=4$ bidders and $N=3$ items, where $B_1=\{(2, 0, 0), 1\}$, $B_2=\{(2, 2, 1), 5\}$, $B_3=\{(0, 1, 1), 2\}$, $B_4=\{(0, 1, 4), 3\}$, and the available units for each item are $u_1=6$, $u_2=3$, and $u_3=4$, respectively.}
	\label{fig:bid_item_graph}
	\vspace{-1em}
\end{figure}

First, we design features for bid nodes. Undoubtedly, the valuation, i.e., the maximum price that each bidder is willing to pay, is an important factor for deciding the optimal allocation. Generally, the auctioneer prefers to satisfy the bidders with higher valuations. However, bidders with higher valuations usually request more units than the ones with lower valuations. Therefore, the number of total required units in each bid also has an influence on the final allocation. To capture this, we choose the valuation, $p_m$, and the total number of required units, $\sum_{n=1}^N \lambda_m^n$, as two features for the node $v_{b_m}$. Therefore, the feature vector of $v_{b_m}$ is defined as $\bm{x}_{b_m}=[p_m, \sum_{n=1}^N \lambda_m^n] $. As for the item nodes, we design two features to demonstrate their characteristics. Specifically, for the node $v_{\iota_n}$, the first feature is the number of its available units, $u_n$, which is directly related to the constraint (\ref{WDP_sub1}). And the second feature is the popularity index $d_n$,  which is defined as the number of bidders that request $\iota_n$ and corresponds to the degree of $v_{\iota_n}$. In this way, the feature vector of $v_{\iota_n}$ is given by $\bm{x}_{\iota_n}=[u_n, d_n]$.  Finally, as mentioned earlier, we define the feature vector of the edge $e(b_m,\iota_n)$, if it exists, as $\bm{e}(b_m,\iota_n)=[\lambda_m^n]$. Based on the aforementioned procedure, we include all the information in Problem (\ref{WDP}) into the augmented bid-item graph, an example of which can be found in Fig. \ref{fig:bid_item_graph}. In the following, we introduce the specific GNN structure for this augmented  bipartite bid-item graph.

\subsection{GNN Structure for Multi-Unit WDP}\label{s3_3}
\begin{figure*}
	\centering
	\includegraphics[width=0.9\linewidth, height=0.25\textheight]{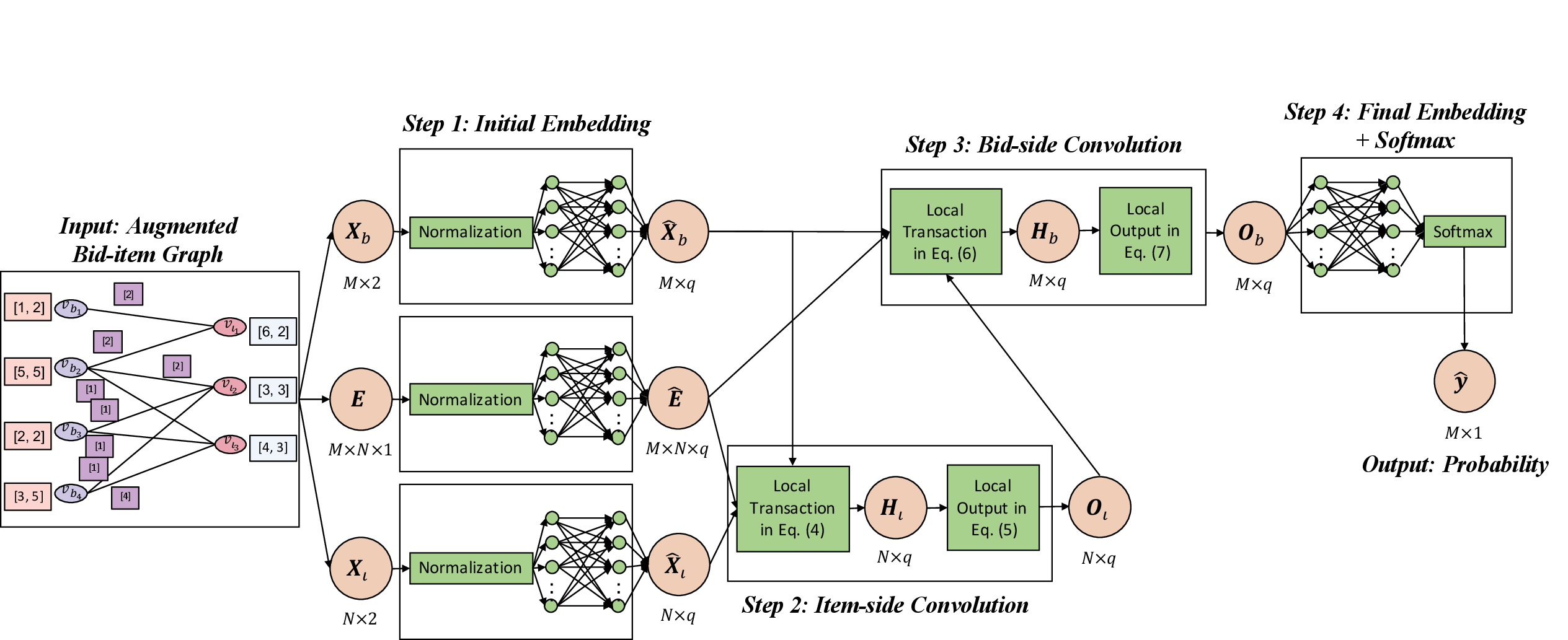}
	\caption{Structure of the GNN with half-convolution operations, which comprises an initial embedding process, two successive half-convolution operations, and a final embedding process. It takes the augmented bid-item graph as input and outputs a probability map over all the bid nodes. The bold upper-case letters indicate the matrices by row stacking the vectors represented as the corresponding bold lower-case letters. For example, $\bm{X}_b = [\bm{x}_{b_1};\bm{x}_{b_2};...;\bm{x}_{b_M}]$ is an $M\times 2$ matrix, whose $m$-th row is equal to $\bm{x}_{b_m}$. }
	\label{fig:bi_gnn}
	\vspace{-1em}
\end{figure*}

In this section, we introduce an approach to learn the probability of each bid node belonging to the optimal allocation. Given that the augmented bid-item graph is  bipartite, we adopt the half-convolution operation \cite{bi_gnn}  to achieve the above goal. Half-convolution operation is especially designed for the bipartite graph. It breaks down the local transition and output functions mentioned in Section \ref{s3_1} into two successive passes starting from the item nodes. In this way, we regard the outputs of item nodes as intermediate results and mainly focus on the predicted probability of bid nodes. It is different from the general GNN model in Section \ref{s3_1} that aims to learn the outputs of all the nodes, and thus reduces the training complexity. The structure of the GNN with half-convolution operations is summarized in Fig. \ref{fig:bi_gnn}.

Specifically, for our bid-item graph, $G(\mathcal{V}_b, \mathcal{V}_\iota, \mathcal{E})$, the features of the bid nodes, the item nodes, and the edges are first normalized and then embedded into different $q$-dimensional vectors. As mentioned in Section \ref{s3_1}, the embedding dimension, $q$, is a hyperparameter and its value is generally manually set. Moreover, the above embedding process is achieved by using 2-layer fully-connected networks and aims to unify the dimensions of node and edge features for the following half-convolution operations. This is presented as Step 1 in Fig. \ref{fig:bi_gnn}. Then we perform the half-convolution operation for each item node in $\mathcal{V}_\iota$ as follows:
\begin{equation}
\bm{h}_{\iota_n}= \sum_{v_{b_m} \in ne[v_{\iota_n}]} g_\iota(\bm{\hat{x}}_{b_m}, \bm{\hat{x}}_{\iota_n}, \bm{\hat{e}}(b_m,\iota_n)), \label{tran_1}
\end{equation}
\begin{equation}
\bm{o}_{\iota_n} = r_\iota(\bm{\hat{x}}_{\iota_n},\bm{h}_{\iota_n}), \label{out_1}
\end{equation}
where $ne[v_{\iota_n}]$ is the set of neighbors of node $v_{\iota_n}$,  which corresponds to the bid nodes that request $\iota _n$. Also, $\bm{\hat{x}}_{b_m}$, $\bm{\hat{x}}_{\iota_n}$, and $\bm{\hat{e}}(b_m,\iota_n)$ are the initial embedding results for the features of node $v_{b_m}$, node $v_{\iota_n}$, and edge $e(b_m,\iota_n)$, respectively. Moreover, functions $g_\iota$ and $r_\iota$ are shared by all item nodes and each function is realized through  a 2-layer fully-connected network with ReLU activation functions. By the operations in (\ref{tran_1}) and (\ref{out_1}), the output of each item node contains the information from the bids that request it. After that, the half-convolution operation is applied to each bid node in $\mathcal{V}_b$ as follows:
\begin{equation}
\bm{h}_{b_m}= \sum_{v_{\iota_n} \in ne[v_{b_m}]} g_b(\bm{o}_{\iota_n}, \bm{\hat{x}}_{b_m}, \bm{\hat{e}}(b_m,\iota_n)), \label{tran_2}
\end{equation}
\begin{equation}
\bm{o}_{b_m} = r_b(\bm{\hat{x}}_{b_m},\bm{h}_{b_m}), \label{out_2}
\end{equation}
where functions $g_b$ and $r_b$ are shared among all the bid nodes and each of them is again realized through a 2-layer fully-connected network with ReLU activation functions. Similarly, by the operations in (\ref{tran_2}) and (\ref{out_2}), the output of each bid node contains the information of the item nodes in its requested bundle and the bid nodes that have at least one item in common with it (which is encapsulated in $\{\bm{o}_{\iota_n}\}_{v_{\iota_n} \in ne[v_{b_m}]}$). The above information is what is needed for computing  the probability of each bid belonging to the optimal allocation. Note that equations (\ref{tran_1}) and (\ref{tran_2}) correspond to the transition function in (\ref{local_tran}), while  equations (\ref{out_1}) and (\ref{out_2}) correspond to the output function in (\ref{local_out}). Specifically, equations (\ref{local_tran}) and (\ref{local_out}) are general expressions while equations  (\ref{tran_1})-(\ref{out_2}) are the detailed realizations we adopt for the WDP. Finally, we apply another 2-layer fully-connected network for the outputs of bid nodes and use the softmax operation to compute a probability distribution over all the bids, which is denoted as $\bm{\hat{y}}$.\footnote{Fig. \ref{fig:bi_gnn} is a centralized implementation of the GNN with half-convolution operations. Each function in (\ref{tran_1})-(\ref{out_2}) is realized through a 2-layer fully-connected network and used by all the nodes. Equivalently, one may assume that each node has a copy of transition and output functions in (\ref{tran_1})-(\ref{out_2}). Using GNN to compute the output of each node is equivalent to the process that each node collects neighborhood information and uses its copy of transition and output functions to compute its own output. }

\subsection{Sample Generation for Multi-Unit WDP}\label{s3_4}
Now that we have introduced the GNN structure for the augmented bid-item graph of the multi-unit WDP, we develop an approach to generate training samples for the GNN to learn over. Generally, a GNN model can be trained  either in supervised or unsupervised manner. However, the unsupervised manner is known as particularly challenging \cite{guided_tree}. Therefore, we train the GNN in Fig. \ref{fig:bi_gnn} using the supervised manner, where the ground truth, i.e., the label, for each training sample is needed. To this end, we first generate $L$ multi-unit WDP instances by following some specific distributions, such as the decay distribution in \cite{data_distribution,data_distribution_2}. Then we obtain the optimal allocations for each instance by using some well-recognized solvers, such as CPLEX. In this way, we obtain an instance set $\mathcal{Q}=\{G^{(l)},\bm{a}^{(l)}\}_{l=1}^L$, where $G^{(l)}$ is the augmented  bid-item graph with features for the multi-unit WDP and $\bm{a}^{(l)}$ is the corresponding optimal allocation result. 

Note that $\bm{a}^{(l)}$ is a binary vector where one appears more than once. If we directly regard $G^{(l)}$ as the input and $\bm{a}^{(l)}$ as its corresponding label, using the GNN to learn $\bm{a}^{(l)}$ is equivalent to a multi-label classification task. However, it is generally difficult to deal with the multi-label classification task, because the size of its output space, i.e., the number of label sets, grows exponentially with the number of classes \cite{multi_label}. To overcome this challenge, different strategies have been proposed to exploit the label correlations and facilitate the learning process \cite{multi_label2}. In this paper, we adopt a first-order strategy  \cite{multi_label3}, as its computational efficiency can enhance the sample generation process and decrease the number of needed label instances. Specifically, we ignore the label correlations and decompose the multi-label classification task into many independent single-label classification tasks. In the following, we introduce detailed procedures to obtain single-label samples for the multi-unit WDP.

\subsubsection{Optimum-Only Sample Generation}\label{s3_4_1}
\begin{figure}
	\centering
	\includegraphics[width=1\linewidth, height=0.25\textheight]{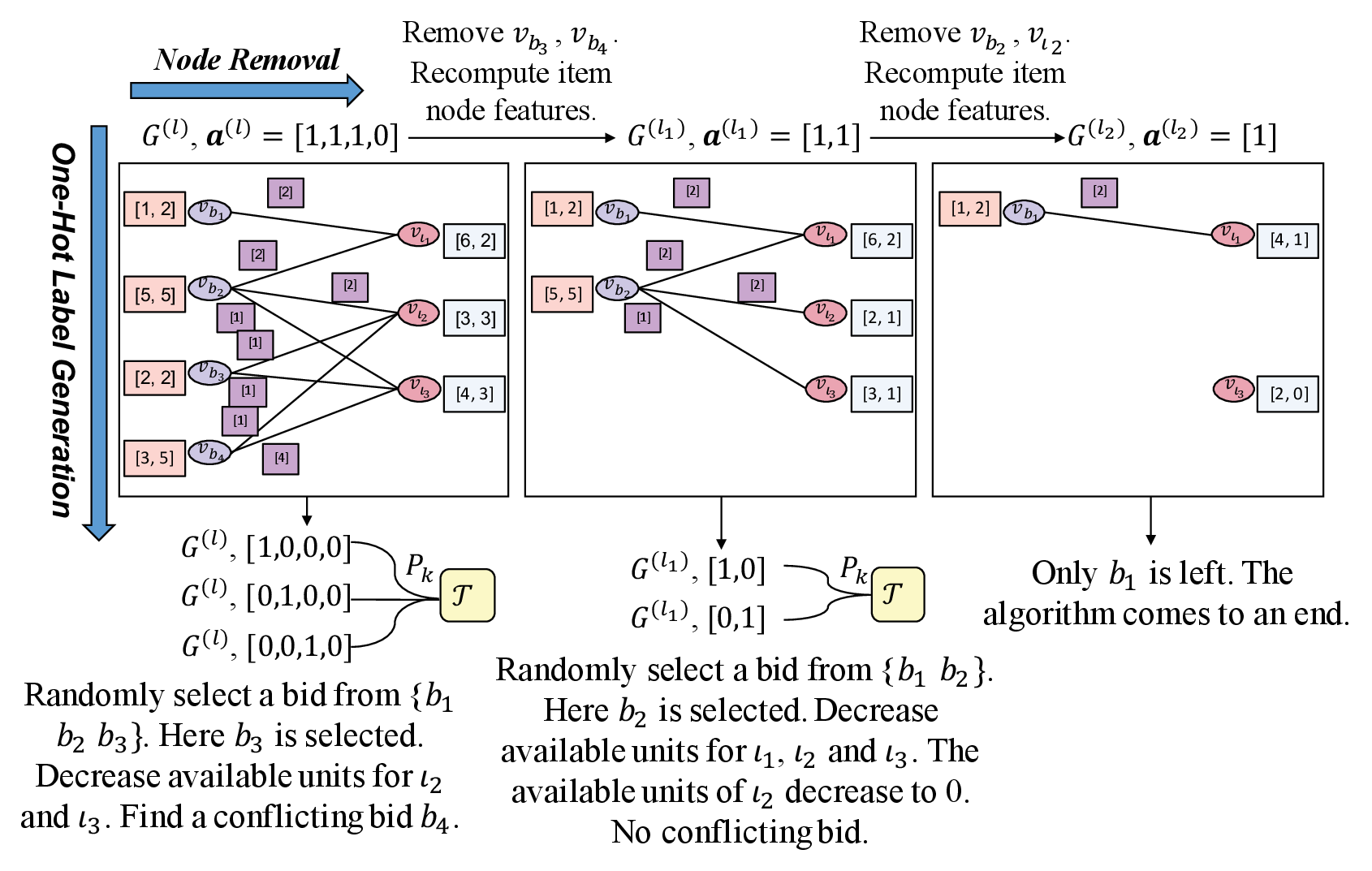}
	\vspace{-2em}
	\caption{Single-label sample generation process for multi-unit WDP described by the augmented bid-item graph in Fig. \ref{fig:bid_item_graph}.  The samples are generated by iteratively running the one-hot label generation algorithm and the node removal process.}
	\label{fig:sample_generation}
	\vspace{-1em}
\end{figure}

To obtain single-label samples from an instance set $\mathcal{Q}$, we propose a \emph{single-label sample generation algorithm} (see Algorithm \ref{sample1}).  For each instance, $\{G^{(l)},\bm{a}^{(l)}\}$, in $\mathcal{Q}$, we first run the \emph{one-hot label generation algorithm} (see Algorithm \ref{onehot}). Specifically, we check each bit in $\bm{a}^{(l)}$. If the $j$-th bit in $\bm{a}^{(l)}$ is equal to 1, a training sample $\{G^{(l)},\bm{y}^{(l_j)}\}$ is generated, where $\bm{y}^{(l_j)}$ is a one-hot label whose $j$-th bit is 1 and all other bits are set to 0. The newly generated sample is added to the training dataset, $\mathcal{T}$, with node keeping probability, $P_k \in [0,1]$. For example, the optimal allocation of the multi-WDP instance in Fig. \ref{fig:bid_item_graph} is [1,1,1,0].  By running Algorithm \ref{onehot}, we obtain three one-hot label samples for this instance, which are labeled as  [1,0,0,0], [0,1,0,0], and [0,0,1,0], respectively. Then the generated samples are added into $\mathcal{T}$ with probability $P_k$. After running the one-hot label generation algorithm, we conduct a \emph{node removal process}. To this end,  we first randomly choose an allocated bid and remove its corresponding node from $G^{(l)}$. For each item allocated to the selected bid, we deduct the allocated units from the item's available units. Note that if the remaining available units for any item decrease to 0, the corresponding item node will also be removed. Then, we traverse the remaining bids to remove all the conflicting bids, i.e., those bids whose requested bundle cannot be satisfied by the remaining available items. Finally, we recompute the features for each item node and update the optimal allocation by removing the corresponding bits for the selected bid and the conflicting bids from $\bm{a}^{(l)}$. Through the above node removal process, we obtain a new graph $G^{(l_1)}$ and its corresponding new allocation vector $\bm{a}^{(l_1)}$. We repeat the one-hot label generation algorithm and the node removal process until only one bid node is left in the graph. An example of the above single-label sample generation process is illustrated in Fig. \ref{fig:sample_generation}. 
\begin{algorithm}
	\caption{Single-Label Sample Generation}
	\label{sample1}
	{\scriptsize
		\begin{algorithmic}[1]
			\STATE \textbf{initialization}
			\STATE \quad Set training dataset: $\mathcal{T}=\emptyset$.
			\FOR {problem instance $\{G^{(l)},\bm{a}^{(l)}\}$ \textbf{in}  $\mathcal{Q}$}
			\STATE $G^{(l_0)}=G^{(l)}$, $\bm{a}^{(l_0)}=\bm{a}^{(l)}$, $s=0$.
			\WHILE {$G^{(l_{s})}$ has more than one bid nodes}
			\STATE $\mathcal{T}_{new}$ = 	One-HotLabelGeneration($G^{(l_s)},\bm{a}^{(l_s)},P_k$).
			\STATE $ \mathcal{T} \leftarrow \mathcal{T} \cup \mathcal{T}_{new}$.
			\STATE Run node removal process to get a new instance $\{G^{(l_{s+1})},\bm{a}^{(l_{s+1})}\}$.
			\STATE $s=s+1$.
			\ENDWHILE
			\ENDFOR
			\RETURN $\mathcal{T}$
		\end{algorithmic}}
\end{algorithm}

\begin{algorithm}
	\caption{One-HotLabelGeneration($G,\bm{a},P_k$)}
	\label{onehot}
	{\scriptsize
		\begin{algorithmic}[1]
			\STATE \textbf{initialization}
			\STATE \quad Set dataset: $\mathcal{T}=\emptyset$.
			\STATE \quad Set index: $j=1$.
			\WHILE {$j \leq length(\bm{a})$}
			\IF {$a_j=1$}
			\STATE Set $\bm{y}$ as a zero vector that has the same length of $\bm{a}$.
			\STATE Set the $j$-th bit of $\bm{y}$ as 1.
			\STATE With probability $P_k$, $ \mathcal{T} \leftarrow \mathcal{T} \cup \{G,\bm{y}\}$.
			\ENDIF
			\STATE $j=j+1$.
			\ENDWHILE
			\RETURN $\mathcal{T}$
	\end{algorithmic}}
\end{algorithm}

The proposed sample generation algorithm is efficient in terms of quantity and variety. On the one hand, with the help of the one-hot label generation algorithm and the node removal process, Algorithm \ref{sample1} can generate $P_k(\sum_{m=1}^{M} a^{(l)}_m-1)(\sum_{m=1}^{M} a^{(l)}_m+2)/2$ training samples for each instance, $\{G^{(l)},\bm{a}^{(l)}\}$, in $\mathcal{Q}$. Therefore, we obtain a large number of training samples from a relatively small instance set. On the other hand, the generated samples are of different sizes because of the node removal process. This enhances the variety of the training samples and the generalization ability of the learned GNN model in terms of problem size.

Note that we can adjust the dataset variety and the sample generation efficiency by choosing different values for node keeping probability, $P_k$. Specifically, a larger $P_k$ leads to a higher sample generation efficiency but lower dataset variety.  We can choose an appropriate $P_k$ according to the specific requirements on the training dataset in practice. Given that each instance in $\mathcal{Q}$ is labeled by its corresponding optimal allocation result, using Algorithm \ref{sample1} to generate samples from $\mathcal{Q}$  is dubbed the \emph{optimum-only sample generation process}.

\subsubsection{Mix Sample Generation}\label{s3_4_2}
Given that the labeled training instances are hard to acquire in practice, we propose an \emph{instance set expansion process} to further improve the sample generation efficiency and decrease the number of needed labeled instances. The expansion process is conducted before running Algorithm \ref{sample1}. Specifically, for each instance, $\{G^{(l)},\bm{a}^{(l)}\}$, in $\mathcal{Q}$, we do a rapid \emph{local search} around the optimal allocation, $\bm{a}^{(l)}$, to get some sub-optimal allocations by simply swapping, deleting, and inserting allocated bids such that the achieved revenue has a small gap to the optimal one, say less than $1\%$.  Note that local search is a widely-used technique for WDPs and other NP-hard problems to adjust solutions, such as the MIS problem \cite{local1} and the maximum clique problem \cite{local2}. Then we include $G^{(l)}$ with each obtained sub-optimal allocation into $\mathcal{Q}$ to obtain an expanded instance set $\mathcal{\hat{Q}}$. Finally, we run the proposed single-label sample generation algorithm, i.e. Algorithm \ref{sample1}, on  $\mathcal{\hat{Q}}$ to generate training samples.

Because instances in $\mathcal{\hat{Q}}$ are labeled by optimal or sub-optimal allocation results, generating samples from $\mathcal{\hat{Q}}$ is dubbed the \emph{mix sample generation process}. Obviously, the mix sample generation is more efficient than the optimum-only one and is more suitable for scenarios where the number of labeled instances is small. However, it inevitably leads to some performance loss because it adds some sub-optimal solutions into $\mathcal{T}$. In Section \ref{s4}, we will further discuss the advantages and disadvantages for each sample generation process based on simulation results.

\subsection{Training Process of the GNN for Multi-Unit WDP} \label{s3_5}
By using the above sample generation processes, we obtain a single-label training sample set, $\mathcal{T}$. Then we train the GNN in Fig. \ref{fig:bi_gnn} with $\mathcal{T}$ in a supervised manner. Given that the training process is similar to that of a binary classifier, we utilize cross entropy as the loss function and the gradient descent algorithm to update the parameters in the GNN.  Because the loss function is related to the misclassification rate, minimizing it  intuitively enables the GNN to solve  the classification problem and predict the probability of each bid belonging to the optimal allocation.

\subsection{Post-Processing Algorithms for Multi-Unit WDP} \label{s3_6}
After the training process in Section \ref{s3_5}, we obtain a GNN model, whose output is a probability map in $[0,1]^M$ that indicates the probability of each bid belonging to the optimal allocation. To transform the continuous probability map into a feasible binary allocation result of the multi-unit WDP in (\ref{WDP}), we propose two novel graph-based post-processing algorithms in the following.
\subsubsection{Basic Post-Processing Algorithm} \label{s3_6_1}
\begin{figure}
	\centering
	\includegraphics[width=1\linewidth, height=0.18\textheight]{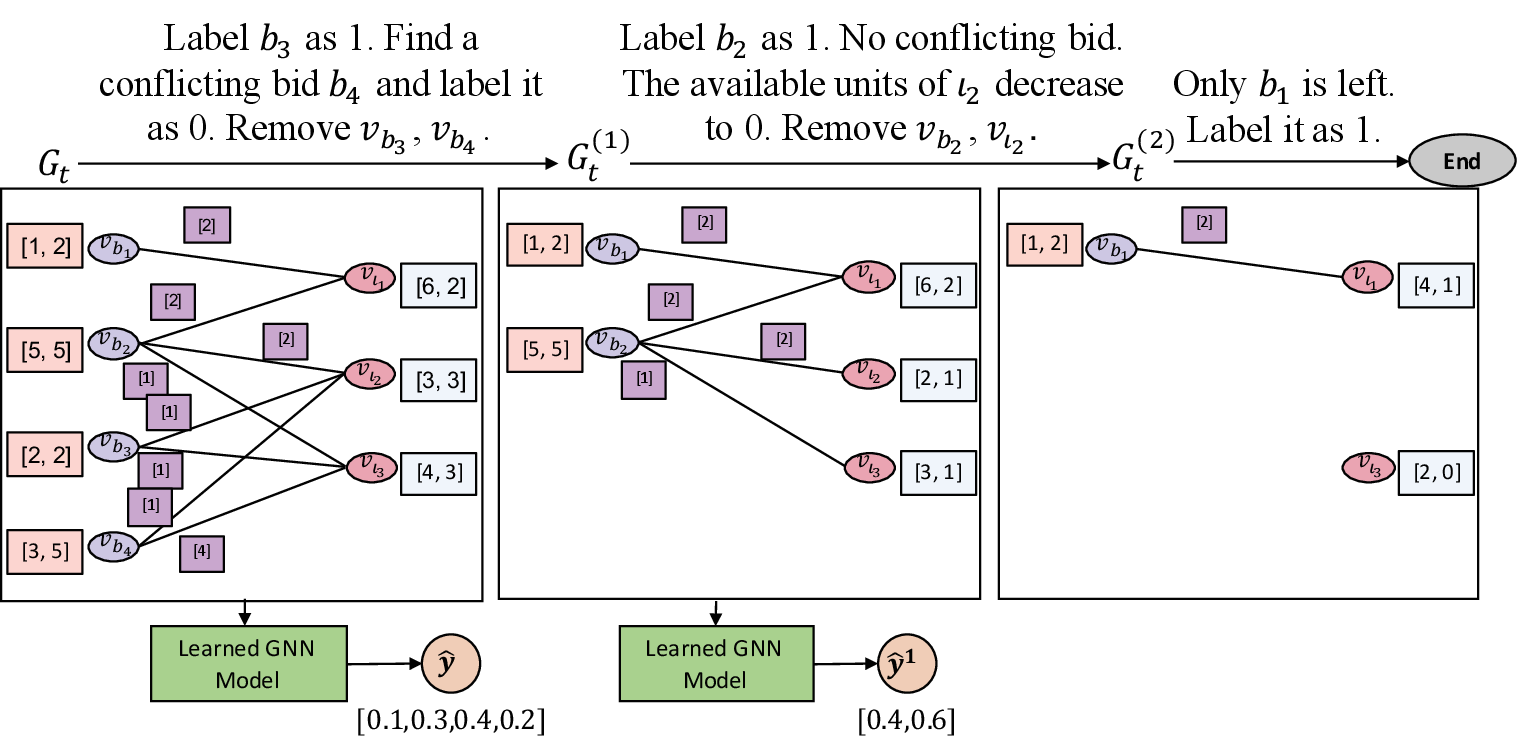}
	\caption{Basic GNN-based method for multi-unit WDP  described by the augmented bid-item graph in Fig. \ref{fig:bid_item_graph}. At each iteration, only the bid node with the highest probability is labeled as 1.}
	\label{fig:basic}
	\vspace{-1em}
\end{figure}

\emph{Basic post-processing algorithm} is similar to the single-label sample generation algorithm, i.e. Algorithm \ref{sample1}. Specifically, for a multi-unit WDP and its corresponding graph, $G_t$, we first use the learned GNN model to get the predicted probability for each bid and  sort all the unlabeled bids in a descending order based on their predicted probabilities. Then we label the bid with the highest rank, i.e., the highest probability, as 1. For each item allocated to the bid, we deduct the allocated units from the item's available units. After that, we traverse the remaining bids in the sorted list and label the conflicting bids as 0. Note that as mentioned in Section \ref{s3_4_1}, conflicting bids are the bids whose bundle cannot be satisfied by the remaining available items. We remove all the labeled bid nodes, recompute the item nodes' features, and remove the item nodes with no available unit from $G_t$ to obtain a residual graph $G_t^{(1)}$, which is then fed into the GNN model to get a new probability map. We repeat the above process until all the bids are labeled. The GNN-based method with the basic post-processing algorithm is dubbed the \emph{basic GNN-based method}, an example of which is illustrated in Fig. \ref{fig:basic}.

As shown in Fig. \ref{fig:basic}., only one bid is assigned as 1 at each iteration. However, we need to find out all the allocated bids and label them as 1. Therefore, the required iteration number of the basic post-processing algorithm is equal to the number of allocated bids, which is positively related to the percentage of satisfied bidders/users, i.e., user satisfaction. Generally, auctions with abundant available units have high user satisfactions. Under this circumstance, the basic post-processing algorithm is time-consuming and not an appropriate choice.

\subsubsection{Traversal Post-Processing Algorithm} \label{s3_6_2}
\begin{figure}
	\centering
	\includegraphics[width=1\linewidth, height=0.18\textheight]{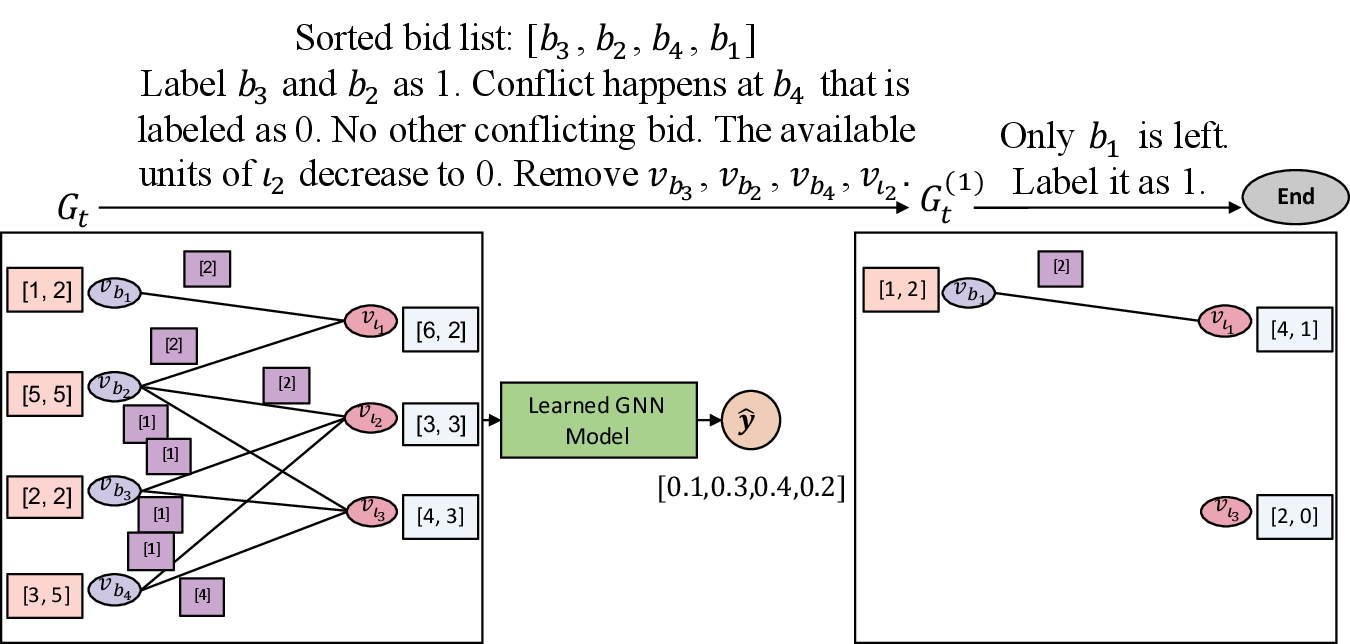}
	\caption{Traversal GNN-based method for multi-unit WDP described by the augmented bid-item graph in Fig. \ref{fig:bid_item_graph}. Compared with the basic GNN-based method, more bids are labeled at each iteration and fewer iterations are needed.}
	\label{fig:traversal}
	\vspace{-1em}
\end{figure}

To accelerate the basic post-processing algorithm mentioned above, we propose a \emph{traversal post-processing algorithm} in the following. Similar to the basic post-processing algorithm, we first use the learned GNN model to get the predicted probability for each bid. Then, we sort all the unlabeled bids in a descending order based on their predicted probabilities. After that, we traverse the bids in the sorted order. For each bid, if its requested bundle can be satisfied by current available items, we label it as 1. And for each item allocated to the bid, we deduct the allocated units from the item's available units and move to the next bid in the sorted list. Otherwise, we label the bid as 0, traverse the remaining bids, and find out the conflicting bids to label them as 0. We remove all the labeled bid nodes, recompute the item nodes' features, and remove the item nodes with no available unit to get a residual graph for the next iteration. Similar to the basic post-processing algorithm, the traversal post-processing algorithm comes to a stop when all the bids are labeled. The GNN-based method with the traversal post-processing algorithm is dubbed the \emph{traversal GNN-based method}, an example of which is illustrated in Fig. \ref{fig:traversal}.

As compared to the basic post-processing algorithm, more bids are labeled at each iteration of the traversal post-processing algorithm. Therefore, the traversal post-processing algorithm  needs fewer iterations and is faster than the basic post-processing algorithm. On the other hand, the traversal post-processing algorithm may lead to some revenue loss, the reason of which is explained as follows. In the basic post-processing algorithm, the GNN is supposed to accurately find out the most probable bid to be allocated. On the contrary, the predicted probability map over all bids is supposed to be accurate in the traversal post-processing algorithm. Therefore, the traversal post-processing algorithm imposes higher requirements on the learned GNN model and may lead to revenue loss. Overall, there is a trade-off between the time complexity and the revenue loss, which will be further discussed in Section \ref{s4}.

\subsection{Time Complexity Analysis} \label{s3_7}
In this section, we analyze the time complexity of the proposed GNN-based methods  as functions of the number of bids, $M$, and the number of items, $N$.
\begin{itemize}[leftmargin=3mm]
	\item \textbf{Basic GNN-based method}: As mentioned in Section \ref{s3_6_1}, the required iteration number of the basic post-processing algorithm  is equal to the number of allocated bids, denoted as $M_b$. Note that $M_b \leq M$ and is determined by $M$, $N$, and other settings in the multi-unit CA. At each iteration of the basic post-processing algorithm,  we first need to run the inference process of the GNN in Fig. \ref{fig:bi_gnn}, where the most time-consuming part is the two half-convolution operations. As shown in (\ref{tran_1}), we need to collect the neighboring bid nodes' information for each item node during the item-side convolution, whose required time is linear in $MN$. Similar analysis can be conducted for the bid-side convolution. Therefore, the complexity of the GNN inference process is $O(MN)$. After that, we sort all the bids, select the bid with the highest probability,  and check all the remaining bid, for which the required time is $O(M\log M)$. The overall time complexity of one iteration in the basic post-processing is $O(MN+M\log M)$. Given that $M_b$ iterations are needed, the time complexity of the basic GNN-based method is $O(M_b(MN+M\log M))$. 
	
	\item \textbf{Traversal GNN-based method}: As mentioned in Section \ref{s3_6_2}, the traversal post-processing needs fewer iterations than the basic one. We denote the  required iteration numbers  for the traversal post-processing as $M_t$. Note that $M_t \leq M_b \leq M$. Given that the operations at each iteration of the traversal post-processing algorithm are similar to those in the basic one, the time complexity of the traversal GNN-based method is given by $O(M_t(MN+M\log M))$. 
\end{itemize}
From the above analysis, the traversal GNN-based method generally enjoys a lower time complexity than the basic GNN-based method. The conclusion is consistent with the analysis in Section \ref{s3_6_2} and will be further verified through the simulation results in Sections \ref{s4} and \ref{s5}.

\section{Synthetic Testing Results} \label{s4}
In this section, we conduct extensive simulations on the synthetic instances to verify the effectiveness of our proposed GNN-based method. 

\subsection{Simulation Setup} \label{s4_1}
\subsubsection{Instance Generation} \label{s4_1_1}
Because few real-world data exists for the multi-unit WDP, existing works usually generate synthetic instances from manually designed distributions. In this section, we adopt the distribution considered in \cite{data_distribution,data_distribution_2} to obtain multi-unit WDP instances for training, validation, and testing. Following a common practice, we assume that the number of bids is larger than the number of items, i.e., $M>N$. Without loss of generality, we set the ratio between the number of bids, $M$, and the number of items, $N$, i.e., bid-item ratio, as 10. In the rest of this section, we only specify the value of $M$ for simplicity. For each problem instance, we collect $M$ bids according to following steps \cite{data_distribution}.
\begin{itemize}[leftmargin=3mm]
	\item Set the number of units for each item. Specifically, for the $n$-th item, $\iota_n$, we randomly choose the number of units, $u_n$, from $[1, 2, ..., u_{\max}]$, where the maximum number of units, $u_{\max}$, is selected in advance. 
	\item Select the number of items in each bid. This number is drawn from a decay distribution \cite{data_distribution_2}. Specifically, we iterate over all items and add each item into the bid with probability $P_\iota \in [0,1]$ until an item is not added or the bid includes all items. In this section, we set $P_\iota$ as 80\%.
	\item Select the number of units for each item in each bid according to the decay distribution again. Specifically, for each item in a bid, we repeatedly add a unit with probability $P_u \in [0,1]$ until a unit is not added or the bid includes all available units for the item. In this section, we set $P_u$ as 65\%.
	\item Pick the valuation, $p_m$, for each bid. Generally, the valuation of a bid is positively related to the number of units in this bid. Therefore, we randomly choose $p_m$ for each bid  between 0 and the number of units in the bid \cite{data_distribution}.
\end{itemize} 
During the above process, if the same bid is generated twice, the new one will be deleted and regenerated. Moreover, if the newly generated bid is dominated by an existing bid, the new bid will also be deleted. Note that $B_i$ is dominated by $B_j$ if the requested bundle of $B_i$ is a superset of that of $B_j$, and the valuation of $B_i$ is smaller than or equal to that of $B_j$.
In this case, $B_i$ will not belong to the optimal allocation. Similarly, if an existing bid is dominated by the newly generated one, we will delete the existing bid. Removing the bids that are dominated by others is also a widely-used and low-complexity preprocessing method in the existing works for WDPs \cite{CABOB}. In the end, the generated bids are different and not dominated by each other. Only under this circumstance, $M$ and $N$ can be used to accurately evaluate the complexity of the multi-unit WDPs\footnote{The distribution of the generated instances may not be exactly consistent with real-world bidding auctions. The construction of realistic data distributions is application/market dependent, which can be further studied for specific cases in the future.}. 

\subsubsection{Other Parameters}\label{s4_1_2}
We set the dimension of state embedding, $q$, as 16. Therefore, each fully-connected layer of the GNN in Fig. \ref{fig:bi_gnn} consists of 16 neurons. Moreover, following \cite{bi_gnn}, we generate 10,000 random instances for training, 2,000 for validation, and 60 for testing by the instance generation process mentioned above, unless otherwise stated. Then, we set the node keeping probability, $P_k$, as 80\%, and use the sample generation algorithms in Section \ref{s3_4} to collect samples from the training and validation instances, respectively. Specifically, we collect 100,000 training samples and 20,000 validation samples for the subsequent tests in this paper.  Note that the needed numbers of training and validation instances to collect enough samples are smaller than 10,000 and 2,000, respectively. We generate more instances than needed to ensure that enough samples can be collected. 

As for the training process, we minimize the cross entropy loss using the Adam optimizer \cite{adam}, set the batch size as 32 and the initial learning rate as 0.001. We decrease the learning rate by 5 times when the validation loss does not improve for 10 epochs, and stop the training process once it does not improve for 20 epochs. Furthermore, we utilize an existing open-source code for the half-convolution operation\footnote{https://github.com/ds4dm/learn2branch.}, and all other codes are implemented in python 3.6, which run on a Windows system with an Intel i7-9700 3.0 GHz CPU and a 32GB RAM.

\subsection{Results of CPLEX}\label{s4_2}
First, we test the performance of the globally optimal algorithm for the multi-unit WDP, i.e., the branch-and-bound algorithm \cite{bb}, as a benchmark. We use CPLEX, one of the most recognized solver for integer linear programming problems, to implement the branch-and-bound algorithm. Previous works \cite{CABOB} have also used the results of CPLEX as the optimal solutions of the WDP. Given that the complexity of the branch-and-bound algorithm is exponential, we set the time limit for all the tests as 1,800s. For a specific instance, if the CPLEX runs out of time and cannot find its optimal result, its running time is recorded as 1,800s and a sub-optimal solution of this instance is obtained. 
\begin{table}[htbp]
	\vspace{-1em}
	\scriptsize
	\caption{Time for CPLEX}
	\vspace{-1em}
	\label{cplex}
	\centering
	\begin{tabular}{|l|c|c|c|c|c|c|}
		\hline
		\diagbox{$u_{\max}$}{Time(s)}{$M$} &500 &1000 &1500 &2000 &3000& 5000 \\
		\hline
		1 &0.11 &0.41 &0.98 & 1.63& 22.05&1233.07\\
		\hline
		3 &0.17 &0.89 &8.92 &81.36 &1800.00&1800.00\\
		\hline
		5 &0.25 &4.90 &118.62 &1800.00 &1800.00&1800.00\\
		\hline
		8 &0.34 &18.53 &1800.00 &1800.00 &1800.00&1800.00\\
		\hline
		10 &0.63 &49.82&1800.00 &1800.00 &1800.00&1800.00\\
		\hline
	\end{tabular}
\end{table}

The testing results of the CPLEX on 60 testing instances are summarized in Table \ref{cplex}.  From the table, the running time increases with the number of bids and the maximum number of units. It can be seen that when $M\geq1500$ and $u_{\max}\geq8$, CPLEX cannot efficiently find the optimal allocation. Therefore, CPLEX cannot satisfy the latency requirements for large-size WDPs. The results also suggest that collecting training samples from large-size instances is infeasible. Therefore, we can only train the GNN model with small-size instances and use the learned model on large-size instances. For the subsequent tests in this paper, we mainly collect the training dataset from instances with 500 bids that can be solved within 1s to train the GNN model unless otherwise stated.

\subsection{Results of GNN-Based Method }\label{s4_3}
\subsubsection{Convergence Performance and Training Overhead}\label{s4_3_con}
\begin{figure}
	\centering
	\includegraphics[width=0.8\linewidth, height=0.23\textheight]{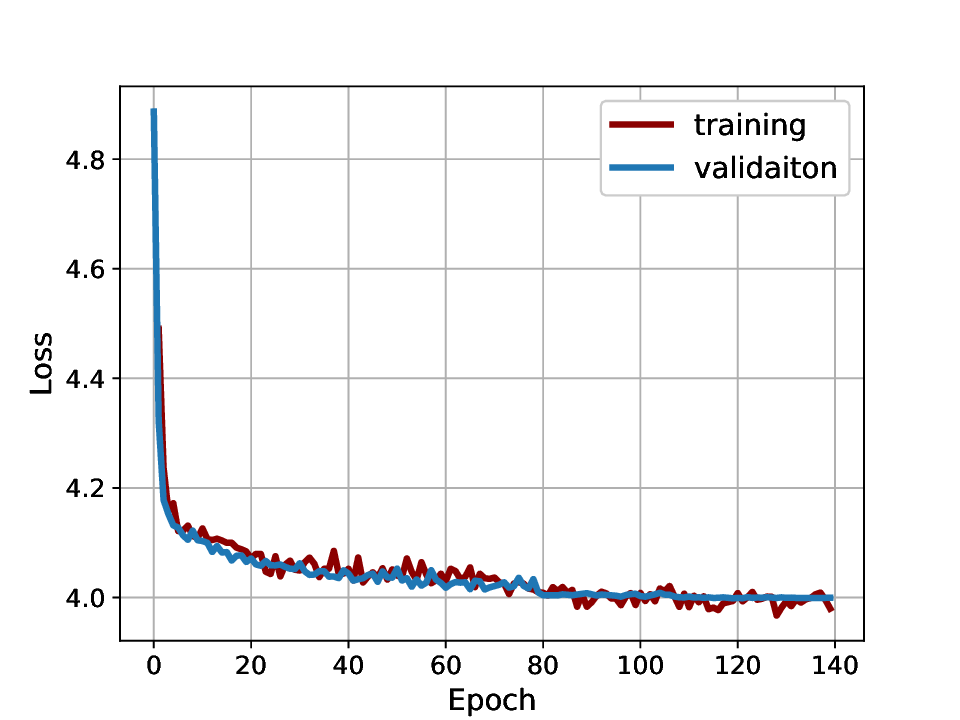}
	\caption{Learning curve of the GNN model in Fig. \ref{fig:bi_gnn}, demonstrating the loss convergence and a small generalization gap.}
	\label{fig:converge}
	\vspace{-1em}
\end{figure}

We first evaluate the convergence performance and the training overhead of the GNN model in Fig. \ref{fig:bi_gnn}. We use the optimum-only sample generation process to collect 100,000 training and 20,000 validation samples from instances with $M = 500$ and $u_{\max} = 5$. The training and validation sample generation processes consume about 16 minutes and 3 minutes, respectively. The resulting training and validation loss over each training epoch is shown in Fig. \ref{fig:converge}.  We see that both training and validation loss converge after about 80 training epochs. Also, the generalization gap, i.e.,  the gap between the training and validation loss, is small, which suggests that the GNN model achieves a good fit.

\subsubsection{Performance of Different Post-Processing Algorithms}\label{s4_3_1}
As mentioned in Section \ref{s3_6}, we have proposed two novel graph-based  post-processing algorithms, i.e., the basic and traversal post-processing algorithms. In this section, we test the performance of the GNN-based method with these two post-processing algorithms, respectively. The training samples are generated from instances with the optimum-only sample generation process in Section \ref{s3_4_1}. The number of bids for the training instances is 500 and the maximum number of units for the training instances is consistent with that of the testing instances. Results are summarized in Tables \ref{table:basic_optimum} and \ref{table:tra_optimum}, where \emph{gap}, also called \emph{performance gap}, means the difference between the revenue achieved by CPLEX and that achieved by the proposed GNN-based method.
\begin{table}[htbp]
	\vspace{-1em}
	\scriptsize
	\caption{Performance of Basic GNN-Based Method with Optimum-Only Sample Generation}
	\vspace{-1em}
	\label{table:basic_optimum}
	\centering
	\begin{tabular}{|l|c|c|c|c|c|}
		\hline
		\diagbox{$u_{\max}$}{Gap and Time(s)}{$M$} &1000 &1500 &2000 &3000 &5000\\
		\hline
		1& \tabincell{c}{4.68\%\\0.27}& \tabincell{c}{4.40\%\\0.38}& \tabincell{c}{4.43\%\\0.55}& \tabincell{c}{4.37\%\\1.11}& \tabincell{c}{3.98\%\\3.77}\\
		\hline
		3& \tabincell{c}{5.67\%\\0.29}& \tabincell{c}{5.10\%\\0.43}& \tabincell{c}{5.26\%\\0.62}& \tabincell{c}{4.47\%\\1.25}& \tabincell{c}{2.88\%\\4.16}\\
		\hline
		5& \tabincell{c}{6.17\%\\0.34}& \tabincell{c}{5.55\%\\0.50}& \tabincell{c}{5.44\%\\0.76}& \tabincell{c}{4.36\%\\1.68}& \tabincell{c}{2.90\%\\5.81}\\
		\hline
		8& \tabincell{c}{6.45\%\\0.37}& \tabincell{c}{6.17\%\\0.62}& \tabincell{c}{5.34\%\\0.95}& \tabincell{c}{3.86\%\\2.21}& \tabincell{c}{2.51\%\\7.87}\\
		\hline
		10& \tabincell{c}{7.25\%\\0.44}& \tabincell{c}{6.59\%\\0.78}& \tabincell{c}{5.02\%\\1.25}& \tabincell{c}{3.82\%\\2.98}& \tabincell{c}{2.66\%\\10.69}\\
		\hline
	\end{tabular}
\end{table}

\begin{table}[htbp]
	\scriptsize
	\caption{Performance of Traversal GNN-Based Method with Optimum-Only Sample Generation}
	\vspace{-1em}
	\label{table:tra_optimum}
	\centering
	\begin{tabular}{|l|c|c|c|c|c|}
		\hline
		\diagbox{$u_{\max}$}{Gap and Time(s)}{$M$} &1000 &1500 &2000 &3000& 5000 \\
		\hline
		1& \tabincell{c}{4.69\%\\0.20}& \tabincell{c}{4.05\%\\0.20}& \tabincell{c}{4.49\%\\0.21}& \tabincell{c}{4.40\%\\0.24}& \tabincell{c}{4.72\%\\0.35}\\
		\hline
		3& \tabincell{c}{5.87\%\\0.20}& \tabincell{c}{5.96\%\\0.21}& \tabincell{c}{5.62\%\\0.22}& \tabincell{c}{4.88\%\\0.27}& \tabincell{c}{3.30\%\\0.38}\\
		\hline
		5& \tabincell{c}{7.63\%\\0.21}& \tabincell{c}{7.10\%\\0.21}& \tabincell{c}{6.41\%\\0.23}& \tabincell{c}{4.87\%\\0.27}& \tabincell{c}{3.69\%\\0.47}\\
		\hline
		8& \tabincell{c}{7.42\%\\0.20}& \tabincell{c}{7.18\%\\0.22}& \tabincell{c}{5.96\%\\0.24}& \tabincell{c}{4.75\%\\0.29}& \tabincell{c}{2.81\%\\0.54}\\
		\hline
		10& \tabincell{c}{8.18\%\\0.21}& \tabincell{c}{7.80\%\\0.22}& \tabincell{c}{6.02\%\\0.25}& \tabincell{c}{5.03\%\\0.32}& \tabincell{c}{3.67\%\\0.63}\\
		\hline
	\end{tabular}
\vspace{-1em}
\end{table}

Compared with Table \ref{cplex}, our proposed method with both post-processing algorithms runs faster than CPLEX. Specifically, for instances with 5,000 bids and the maximum number of units as 10, i.e., the most complicated setting in Table \ref{cplex}, CPLEX cannot solve it within 1,800s. However,  the basic and traversal GNN-based methods can finish within 10.69s and 0.63s, respectively with small performance loss. Moreover, the training samples for all  tests in Tables \ref{table:basic_optimum} and \ref{table:tra_optimum} are generated from instances with 500 bids while  the numbers of bids for the testing instances are consistently larger than 500.  Given that performance gaps do not increase with the number of bids for the testing instances, our proposed method exhibits a good generalization ability in terms of problem size. Therefore, our model can be trained via small-size instances that can be solved optimally very fast and then be used for large-size problems in practice. In the end, we avoid the training process for large-size problems, which is both time- and memory-consuming. Furthermore, comparing the results in Tables \ref{table:basic_optimum} and \ref{table:tra_optimum}, the traversal GNN-based method runs faster but induces around 1\% larger performance gaps than the basic one, which is consistent with the analysis in Section \ref{s3_6}. This also reveals a  trade-off between the time complexity and the revenue loss. In practice, we will choose the basic post-processing algorithm for revenue-sensitive scenarios and the traversal post-processing algorithm for time-sensitive scenarios.

\subsubsection{Performance of Different Sample Generation Processes}\label{s4_3_2}
As mentioned in Section \ref{s3_4}, we have proposed two different sample generation processes, i.e., the optimum-only and mix sample generation processes. They are similar while the mix one conducts an instance set expansion process before collecting samples. The GNN models with different sample generation processes may obtain different allocation results, which lead to different performance gaps. Furthermore, the running time of the GNN-based method depends on the number of allocated bidders. Therefore, different allocation results also consume different running time.  In this section, we test the running time and performance gap of the GNN-based method with these two sample generation processes. The training samples are still from instances with 500 bids. The maximum number of units is 10 for both the training and testing instances. We adopt the basic post-processing algorithm. The results are summarized in Table \ref{table:sample_generation_compare}.

\begin{table}[htbp]
	\scriptsize
	\caption{Performance of Basic GNN-Based Method with Different Sample Generation Processes}
	\vspace{-1em}
	\label{table:sample_generation_compare}
	\centering
	\begin{tabular}{|l|c|c|c|c|c|}
		\hline
		\diagbox{Sample \\Generation}{Gap and Time(s)}{$M$} &1000 &1500 &2000 &3000 &5000 \\
		\hline
		Optimal-Only& \tabincell{c}{7.25\%\\0.44}& \tabincell{c}{6.59\%\\0.78}& \tabincell{c}{5.02\%\\1.25}& \tabincell{c}{3.82\%\\2.98}& \tabincell{c}{2.66\%\\10.69}\\
		\hline
		Mix & \tabincell{c}{8.33\%\\0.44}& \tabincell{c}{7.60\%\\0.78}& \tabincell{c}{6.48\%\\1.27}& \tabincell{c}{4.99\%\\3.05}& \tabincell{c}{4.70\%\\11.59}\\
		\hline
	\end{tabular}
\end{table}

From Table \ref{table:sample_generation_compare}, the time consumed by the GNN-based method with both sample generation processes is close to each other, while the performance gap of the optimum-only sample generation process is around 1\% smaller than that of the mix sample generation process. On the other hand, to generate the same number of samples from instances with 500 bids, the optimum-only sample generation process needs eight times as many instances as the mix sample generation process, which means that the mix sample generation process is more efficient in terms of sample generation. The testing results are consistent with our analysis in Section \ref{s3_4} and indicate a trade-off between the instance collection overhead and the revenue loss. However, the instance collection stage is generally offline and the overhead may be amortized in practice over different applications. Therefore, we recommend the optimum-only sample generation process unless labeled instances are limited in practice.

\subsection{Comparison with Existing Heuristic Algorithms}\label{s4_4}
In this subsection, we further compare our proposed GNN-based method with several well-known heuristic algorithms in literature. Specifically, we consider three different heuristic algorithms as benchmarks: 
\begin{itemize}[leftmargin=3mm]
	\item \textbf{RLP method  \cite{cloud1}}: In the RLP method, the binary variables, $\{a_1, a_2, ..., a_M\}$, are relaxed into continuous variables between 0 and 1. In this way, Problem (\ref{WDP}) is transformed into a linear programming (LP) problem. By solving this relaxed problem, a continuous allocation result $\hat{a}_m \in [0,1]$ is obtained for each bid and  all bids are sorted in a descending order based on the continuous allocation results. Then we traverse the bids in the sorted order. For each bid, it is temporarily labeled as 1 with probability $\hat{a}_m$. A bid is included in the final allocation if it is temporarily labeled as 1 and it does not violate any constraints in (\ref{WDP_sub1}). 
	\item \textbf{SS method  \cite{SS}}: As mentioned in the RLP method, we obtain a relaxed LP problem by relaxing the binary variables in Problem (\ref{WDP}). In the SS method, we solve the dual problem of the relaxed problem and get the dual price, $\{\hat{p}_1, \hat{p}_2, .. ., \hat{p}_N\}$, for each item. Then we reorder the bids by the decreasing values of $p_m/\sum_{n=1}^{N}\hat{p}_n \lambda_m^n$ and  traverse the bids in the sorted order. We include a bid into the final allocation if it does not violate any constraints in (\ref{WDP_sub1}). 
	\item  \textbf{Casanova \cite{Casanova}}: Casanova is a stochastic local search algorithm. The algorithm starts with an empty allocation vector and successively adds an unallocated bid to the current vector to reach a neighboring  allocation vector.\footnote{In stochastic local search algorithms, two allocation vectors are neighbors if one can be achieved by adding a bid into the other one. } In each step, with walk probability $P_w$, a random bid is chosen. With probability $1-P_w$, a bid is selected greedily by ranking the unallocated bids according to the normalized bid price, $p_m/\sum_{n=1}^{N}\lambda_m^n$. The highest-ranked bid is selected if its \emph{age} is smaller than that of the second bid in the list, where the age of each bid refers to the number of steps since the bid was last added to a candidate solution. Otherwise, we select the highest ranked bid with novelty probability $P_n$, and the second one with probability of $1-P_n$.  In the following test, we set $P_w=0.8$ and $P_n=0.5$. We also use a soft restart strategy following \cite{Casanova}, which reinitializes the search if no improvement in the revenue has been achieved within the last 5,000 steps. The search is restarted 5 times and the best solution from all runs is adopted as the final result\footnote{The parameters for the Casanova algorithm are chosen based on both the simulation results and the parameter settings in  \cite{Casanova}.}. 
\end{itemize}
Among the above three methods, the RLP and the SS methods involve solving LP  and its dual problem, whose worst-case time complexity are exponential. In practice, they can be solved in weakly polynomial time\footnote{A problem can be solved in weakly polynomial time means that its running time depends not only on the number but also on the specific values of the input variables.} \cite{weak}. For example, by using the interior point method, the time complexity of the LP and its dual problem can be roughly represented as $O(M^{3.5}N^2)$ and $O(N^{3.5}M^2)$, respectively. Given that the number of bids is generally larger than the number of items, i.e., $M>N$, the SS method runs faster than the RLP method. However, even using the weakly polynomial algorithms, the time complexity of both methods are still higher than that of the proposed method. As for the Casanova algorithm, the running time is random and influenced by many parameters. Therefore,  the practical time complexity of Casanova is usually considered.

\begin{figure}
	\vspace{-1em}
	\centering
	\subfigure[Gap and time.]{
		\begin{minipage}[t]{0.98\linewidth}
			\centering
			\includegraphics[width=0.8\linewidth]{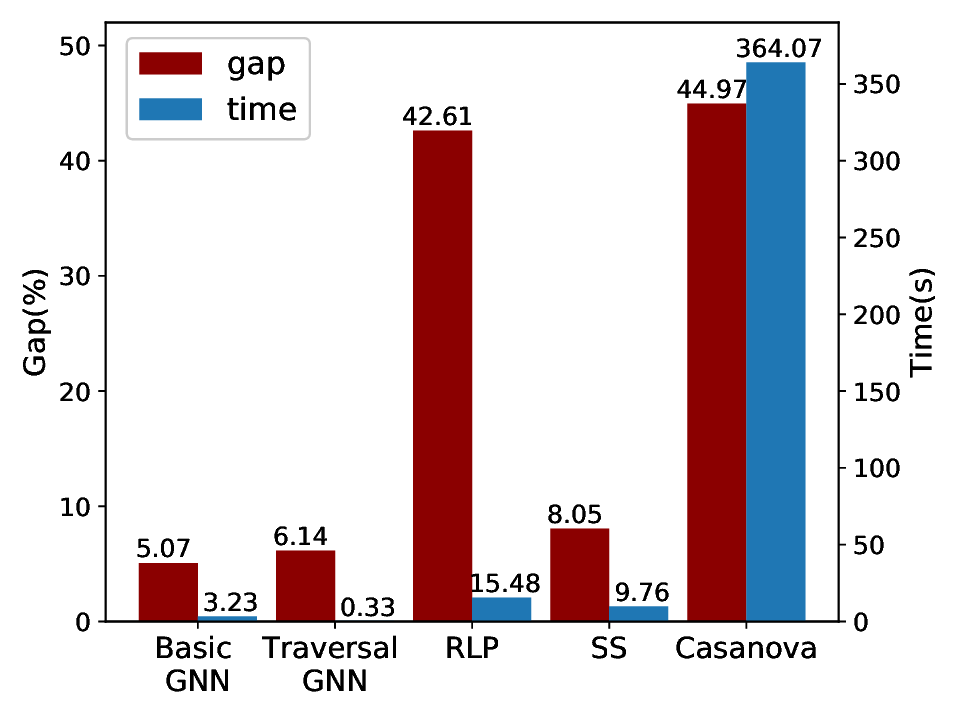}
			\label{general_gap_time}
			\vspace{-1em}
		\end{minipage}
	}
	\subfigure[Utilization and satisfaction.]{
		\begin{minipage}[t]{0.98\linewidth}
			\centering
			\includegraphics[width=0.8\linewidth]{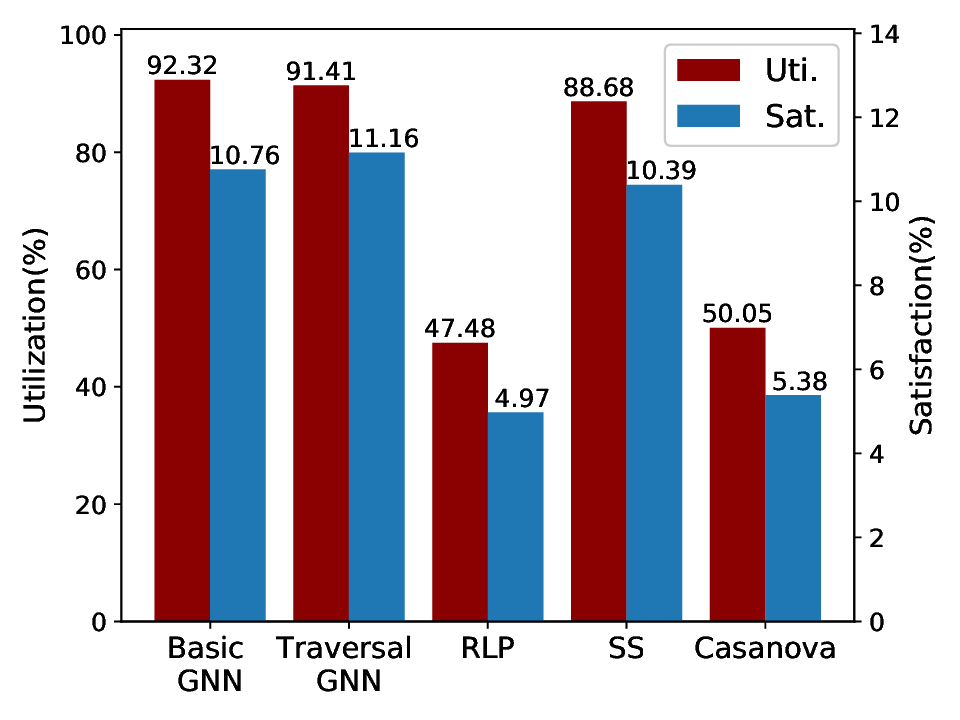}
			\label{general_uti_ser}
			\vspace{-1em}
		\end{minipage}
	}
\vspace{-1em}
	\caption{Average performance of different methods on synthetic instances.}
	\label{fig_general}
	\vspace{-1em}
\end{figure}

In the following, we compare the performance of our proposed method with the above three heuristic algorithms. Specifically, we test the performance gaps and running time of the benchmarks on the instances with different numbers of bids where the maximum number of units is set as 10. Then we compare their performance with our proposed method. We adopt the optimum-only sample generation process for our proposed method. The results are summarized in Table \ref{per_heu}. Moreover, the average performance gaps and running time of different methods over the instances  with different numbers of bidders are plotted in Fig. \ref{general_gap_time}.

\begin{table}[htbp]
	\scriptsize
	\caption{Gap and Time of Different Methods}
	\vspace{-1em}
	\label{per_heu}
	\centering
	\begin{tabular}{|c|c|c|c|c|c|}
		\hline
		\diagbox{Method}{Gap and Time(s)}{$M$}  &1000 &1500 &2000 &3000 &5000\\
		\hline
		Basic GNN& \tabincell{c}{7.25\%\\0.44}& \tabincell{c}{6.59\%\\0.78}& \tabincell{c}{5.02\%\\1.25}& \tabincell{c}{3.82\%\\2.98}& \tabincell{c}{2.66\%\\10.69}\\
		\hline
		Traversal GNN& \tabincell{c}{8.18\%\\0.21}& \tabincell{c}{7.80\%\\0.22}& \tabincell{c}{6.02\%\\0.25}& \tabincell{c}{5.03\%\\0.32}& \tabincell{c}{3.67\%\\0.63}\\
		\hline
		RLP& \tabincell{c}{43.01\%\\0.74}& \tabincell{c}{42.28\%\\2.28}& \tabincell{c}{42.57\%\\4.68}& \tabincell{c}{43.18\%\\13.61}& \tabincell{c}{42.01\%\\56.10}\\
		\hline
		SS& \tabincell{c}{9.71\%\\0.51}& \tabincell{c}{9.15\%\\1.66}& \tabincell{c}{7.86\%\\3.35}& \tabincell{c}{7.69\%\\9.04}& \tabincell{c}{5.82\%\\34.23}\\
		\hline
		Casanova& \tabincell{c}{41.55\%\\20.18}&
		\tabincell{c}{44.71\%\\98.44}&
		\tabincell{c}{45.09\%\\171.89}&
		\tabincell{c}{45.65\%\\409.48}&
		\tabincell{c}{47.83\%\\1120.38}\\
		\hline
	\end{tabular}
\end{table}

From Table \ref{per_heu} and Fig. \ref{general_gap_time}, it can be seen that our proposed GNN-based method runs faster than the benchmarks. Among these three benchmarks, the SS is faster than the RLP and the Casanova. As for the gaps to the revenues of CPLEX, our proposed method outperforms  the benchmarks for all presented instances. Moreover, the performance gaps induced by the RLP and the Casanova are very large. In fact, they are not suitable for auctions whose available units are limited and user satisfactions are low. Under this circumstance, the RLP method skips bidders that are not temporarily labeled as 1 and a large quantity of items would remain unallocated. As for the Casanova, when the optimal allocation only includes a few bidders, it is very likely that no improvement in the revenue is achieved within a very long time and the searching process stops according  to the soft restart strategy. Therefore, it is difficult for the Casanova to achieve a good revenue within a short time. In Section \ref{s5}, we will further compare the performance of  the proposed method and that of the benchmarks on auctions with abundant available units.
\begin{table}[h]
	\scriptsize
	\caption{Resource Utilization and User Satisfaction of Different Methods}
	\vspace{-1em}
	\label{per_uti_ser}
	\centering
	\begin{tabular}{|l|c|c|c|c|c|}
		\hline
		\diagbox{Method}{Uti. and Sat.}{$M$} &1000 &1500 &2000 &3000 &5000\\
		\hline
		Basic GNN& \tabincell{c}{92.91\%\\10.20\%}& \tabincell{c}{93.58\%\\11.93\%}& \tabincell{c}{92.27\%\\10.70\%}& \tabincell{c}{91.52\%\\10.60\%}& \tabincell{c}{91.31\%\\10.38\%}\\
		\hline
		Traversal GNN& \tabincell{c}{92.00\%\\10.40\%}& \tabincell{c}{91.76\%\\11.93\%}&\tabincell{c}{92.00\%\\11.11\%}& \tabincell{c}{91.03\%\\11.20\%} & \tabincell{c}{90.25\%\\11.14\%}\\
		\hline
		RLP& \tabincell{c}{48.73\%\\4.80\%}& \tabincell{c}{46.67\%\\5.47\%}& \tabincell{c}{49.55\%\\4.95\%}& \tabincell{c}{44.33\%\\4.43\%}& \tabincell{c}{48.11\%\\5.22\%}\\
		\hline
		SS& \tabincell{c}{87.82\%\\9.80\%}& \tabincell{c}{90.79\%\\10.86\%} & \tabincell{c}{88.36\%\\10.70\%}& \tabincell{c}{88.85\%\\9.90\%}& \tabincell{c}{87.60\%\\10.70\%}\\
		\hline
		Casanova& \tabincell{c}{59.09\%\\5.60\%}& \tabincell{c}{52.00\%\\5.80\%}& \tabincell{c}{48.00\%\\4.85\%}&\tabincell{c}{46.55\%\\5.67\%}& \tabincell{c}{44.62\%\\5.00\%}\\
		\hline
	\end{tabular}
\vspace{-1em}
\end{table}

To obtain more insights, we further explore two more metrics to evaluate the proposed method. The first one is resource utilization, which refers to the percentage of the allocated units. The second one is user satisfaction mentioned earlier. We summarize the above two metrics for different methods on instances with the maximum number of units as 10 in Table \ref{per_uti_ser}. Moreover, the average resource utilization and user satisfaction over the testing instances with different numbers of bidders are plotted in Fig. \ref{general_uti_ser}. Compared to the benchmark methods, our proposed method can achieve both higher resource utilization and higher user satisfaction.

\subsection{Influence of the Training Instances' Size} \label{s4_5}
Thus far, we have tested the performance of our proposed method with different post-processing algorithms and sample generation processes, and discussed an approach to choose them in practice. We have also compared our proposed method with some benchmarks to show the advantages of our proposed method. All the above tests have used the training samples generated from instances with 500 bids. In this section, we aim to show how the performance of the proposed GNN-based method changes upon increasing the number of bids for the training instances to 1,000.
\begin{table}[htbp]
	\vspace{-1em}
	\scriptsize
	\caption{Performance of Basic GNN-Based Method Using Training Instances with Different Numbers of Bids}
	\vspace{-1em}
	\label{per_1000_basic}
	\centering
	\begin{tabular}{|l|c|c|c|c|c|}
		\hline
		\diagbox{$M_{train}$}{Gap and Time(s)}{$M_{test}$} &1000 &1500 &2000 &3000 &5000\\
		\hline
		500& \tabincell{c}{7.25\%\\0.44}& \tabincell{c}{6.59\%\\0.78}& \tabincell{c}{5.02\%\\1.25}& \tabincell{c}{3.82\%\\2.98}& \tabincell{c}{2.66\%\\10.69}\\
		\hline
		1000& \tabincell{c}{6.17\%\\0.40}& \tabincell{c}{5.33\%\\0.66}& \tabincell{c}{4.03\%\\0.98}& \tabincell{c}{2.92\%\\2.16}& \tabincell{c}{1.59\%\\7.83}\\
		\hline
	\end{tabular}
\end{table}

We use the optimum-only sample generation process to generate training samples from instances with 1,000 bids and train a new GNN model. Then we test the performance gaps and running time for the basic GNN-based method on instances with different number of bids, where the $u_{\max}$ is set as 10. The results are summarized in Table \ref{per_1000_basic}, where $M_{train}$ and $M_{test}$ refer to the number of bids for training and testing instances, respectively.

From the results, increasing the number of bids for the training instances can further improve the performance of the proposed method in both the performance gap and the running time. Nevertheless, improvement is accompanied by higher overhead for instance collection. As mentioned in Section \ref{s4_2}, the optimal results of large-size instances cannot be efficiently obtained and it is infeasible to collect training samples from large-size instances. This reveals another trade-off between the data collection overhead and performance loss. When implementing the proposed GNN-based method  for a specific scenario in practice, one can first start with small-size training instances and try larger-size training instances if the final performance is not satisfactory.

\subsection{Generalization Ability of GNN-Based Method} \label{s4_6}
In this section, we study the generalization ability of our proposed method. As mentioned in Section \ref{s4_3}, the results in Tables  \ref{table:basic_optimum} and \ref{table:tra_optimum} suggest that the proposed method has a good generalization ability in terms of problem size. In the following, we further discuss the generalization ability of the proposed method on another three aspects.
\subsubsection{Generalization Ability on Maximum Number of Units}\label{s4_6_1}
First, we focus on the generalization ability of our proposed method on the maximum number of units, $u_{\max}$. Different from all the tests mentioned above, $u_{\max}$ for the training and the testing instances are not consistent in the following tests. Specifically, we use the optimum-only sample generation process to collect training samples from instances whose $M$ is 500 and $u_{\max}$ is 5. And the testing instances are with 1,000 bids and different values of $u_{\max}$. We adopt the basic post-processing algorithm and the results are summarized in Tables \ref{gene_um_2} and \ref{gene_um}, where \emph{full training} means using the model trained with instances whose $u_{\max}$ is the same as testing instances, and \emph{generalization} means using the model trained with instances whose $u_{\max}$ is 5.
\begin{table}[htbp]
	\scriptsize
	\caption{Resource Utilization and User Satisfaction for  Basic GNN-Based Method with Different Training Modes}
	\vspace{-1em}
	\label{gene_um_2}
	\centering
	\begin{tabular}{|l|c|c|c|c|c|}
		\hline
		\diagbox{Training Mode}{Uti. and Sat.}{$u_{\max}$} &1 &3 & 5 &8 &10 \\
		\hline
		Full Training& \tabincell{c}{99.00\%\\5.00\%}& \tabincell{c}{94.00\%\\6.70\%}& \tabincell{c}{96.00\%\\7.10\%}&\tabincell{c}{94.89\%\\10.10\%}& \tabincell{c}{92.91\%\\10.40\%}\\
		\hline
		Generalization& \tabincell{c}{97.00\%\\5.90\%}& \tabincell{c}{92.50\%\\6.90\%}& \tabincell{c}{96.00\%\\7.10\%}&\tabincell{c}{92.89\%\\10.70\%}& \tabincell{c}{92.55\%\\10.60\%}\\
		\hline
	\end{tabular}
\end{table}
\vspace{-1em}
\begin{table}[htbp]
	\scriptsize
	\caption{Gap and Time for Basic GNN-Based Method with Different Training Modes}
	\vspace{-1em}
	\label{gene_um}
	\centering
	\begin{tabular}{|l|c|c|c|c|c|}
		\hline
		\diagbox{Training Mode}{Gap and Time(s)}{$u_{\max}$} &1 &3 & 5 &8 &10 \\
		\hline
		Full Training& \tabincell{c}{4.68\%\\0.27}& \tabincell{c}{5.67\%\\0.29}& \tabincell{c}{6.17\%\\0.34}&\tabincell{c}{6.45\%\\0.37}& \tabincell{c}{7.25\%\\0.44}\\
		\hline
		Generalization& \tabincell{c}{9.37\%\\0.29}& \tabincell{c}{9.78\%\\0.30}& \tabincell{c}{6.17\%\\0.34}&\tabincell{c}{6.87\%\\0.41}& \tabincell{c}{8.55\%\\0.45}\\
		\hline
	\end{tabular}
\end{table}

From Table \ref{gene_um_2}, the user satisfaction of the generalization mode is slightly higher than that of the full training mode. As mentioned before, the running time of basic GNN-based method is positively related to the user satisfaction.  Therefore, the running time of the generalization mode is slightly higher than that of the full training one, which is consistent with the results in Table \ref{gene_um}. Moreover,  the achieved resource utilization of the generalization mode is smaller than that of the full training mode, which means more available units are unallocated. This is the reason behind incurring  larger performance gaps using the generalization mode. In the end, the results suggest that the proposed method has limited generalization ability on $u_{\max}$. We need to keep the values of $u_{\max}$ for the training and testing instances close to each other to guarantee good performance.

\subsubsection{Generalization Ability on Bid-Item Ratio} \label{s4_6_2}
In the previous tests, we always keep the bid-item ratio as 10. In this part, we try to test the performance of our GNN-based method when the bid-item ratios for the training and testing instances are inconsistent. Specifically, we use the  optimum-only sample generation process to collect training samples from instances whose $M$ is 500, $N$ is 50, and $u_{\max}$ is 5. Then we test on instances with different numbers of items while keeping their $M$ as 1,000 and  $u_{\max}$ as 5. We adopt the basic post-processing algorithm and the results are summarized in Table \ref{gene_ratio}.
\begin{table}[htbp]
	\vspace{-1em}
	\scriptsize
	\caption{Performance of Basic GNN-Based Method on Instances with Different Bid-Item Ratios}
	\vspace{-1em}
	\label{gene_ratio}
	\centering
	\begin{tabular}{|c|c|c|c|c|c|c|}
		\hline
    	\#Items &50 &100 &150 &200 &300 &500\\
    	\hline
    	\tabincell{c}{Bid-item \\Ratio}&20 & 10&  6.67&  5& 3.33 & 2\\
		\hline
		Gap& 10.78\%&6.17\%&6.55\%&7.19\%&9.43\%&11.16\%\\
		\hline
		Time& 0.31&0.34&0.37&0.43&0.56&0.81\\
		\hline
		Uti.& 96.67\%&96.00\%&90.22\%&85.66\%&78.78\%&65.87\%\\
		\hline
		Sat. & 5.8\%&7.10\%&9.80\%&12.10\%&14.80\%&22.20\%\\
		\hline
	\end{tabular}
\end{table}

From Table \ref{gene_ratio}, with the decrease of bid-item ratio, i.e., the increase of $N$, the user satisfaction undoubtedly increases, which makes the basic GNN-based method become more time-consuming. Moreover, the results from  the 4th to 7th columns of Table \ref{gene_ratio} suggest that the resource utilization is unsatisfactory and many available units remain unallocated when bid-item ratio for the training instances  is larger than that of the testing instances. This can be explained as follows. As mentioned above, the bid-item ratio influences the user satisfaction. A larger bid-item ratio would result in a lower user satisfaction. If a model is learned from instances with a large bid-item ratio, the model tends to allocate items to fewer bidders. If we apply the model to instances with a small bid-item ratio whose user satisfaction is high, the model will allocate items to fewer bidders than the optimal results and many units will remain unallocated.  Furthermore, from the results in the 2nd columns of Table \ref{gene_ratio}, when the bid-item ratio of the training instances is smaller than that of the testing instances, the resource utilization is high but the performance gap is nonetheless large. Similar to the above analysis, a model learned from instances with a small bid-item ratio tends to allocate items to more bidders. When the model is applied to instances with a large bid-item ratio, it may allocate items to excessive bidders. In this way, the resource utilization remains high but the obtained allocation is far away from the optimal one, which leads to a big gap to the optimal revenue. Overall, the results suggest that the proposed method has limited generalization ability on bid-item ratio since instances with different bid-item ratios have different  user satisfactions. We need to keep the values of bid-item ratio for the training and testing instances close to each other to get good performance.

\subsubsection{Generalization Ability on Bid Distribution} \label{s4_6_3}
Up to this point, all the training, validation, and testing instances have been generated from the decay distribution. However, the real-world bidding auctions may follow different distributions. Therefore, in this part, we investigate the influence of the bid distribution on the performance of the proposed method. Specifically, we generate bids according to a new distribution, the uniform distribution \cite{data_distribution_2}, where the number of items for  each bid is fixed as 6. The six items are chosen uniformly at random, while the number of units for each item is randomly chosen between 1 and its available units. We still randomly pick the valuation for each bid between 0 and the number of units in the bid.  Also, we use the optimum-only sample generation process to collect training samples from instances with $M = 500$ and $u_{\max} = 5$, and test on instances with $M = 1,000$ and $u_{\max} = 5$. Adopting the basic post-processing algorithm, the results are summarized in Table \ref{distribution}.

\begin{table}
	\scriptsize
	\caption{Performance of Basic GNN-based Method on Instances from Different Distributions}
	\vspace{-1em}
	\label{distribution}
	\centering
	\begin{tabular}{|l|c|c|}
		\hline
		\diagbox{Training Distribution}{Gap and Time(s)}{Testing Distribution}&Decay&Uniform \\
		\hline
		Decay&  \tabincell{c}{6.17\%\\0.34}&  \tabincell{c}{11.88\%\\0.21}\\
		\hline
		Uniform&  \tabincell{c}{17.57\%\\0.46}& \tabincell{c}{7.08\%\\0.21} \\
		\hline
	\end{tabular}
\vspace{-2em}
\end{table}

From Table \ref{distribution}, we see that if the distributions of the training and testing instances are consistent, the performance of the proposed method is similar for different distributions. Otherwise, the performance of the proposed method  deteriorates. Therefore, when using the proposed method to solve WDPs in practice, we need to collect training samples from the real-word bidding auctions we are optimizing, such as from past observations or logging files. In this way, it is expected that the performance of the proposed method will not be severely affected.

\vspace{2mm}
In summary, our method has good generalization ability on the problem size. However, it has limited generalization ability on the maximum number of units, the bid-item ratio, and the bid distribution. Therefore, estimating theses three parameters can significantly boost the performance of our proposed method. In Section \ref{s5_3}, we will further investigate the generalization ability of our method on distributions for different types of cloud computing users.

\section{Testing Results on VM Allocation Problem} \label{s5}
In this section, we focus on a specific multi-unit WDP in the cloud computing, i.e.,  the VM allocation problem, to further validate the performance of the proposed method.

\subsection{VM Allocation In Cloud Computing and Instance Generation}\label{s5_1}
Generally, the cloud providers (such as Amazon and Google) commercialize their abundant computing resources to individuals and enterprises through the cloud computing platform. A cloud computing platform can abstract the physical resources from the users and provide them with the view of VMs. Efficiently allocating VM to users is important in the cloud computing system. Many existing works formulate it as the WDP and then develop different allocation algorithms to solve it \cite{cloud1,cloud2}. 

In the following test, we assume that there are 90 types of VMs and each of them has 500 units according to Amazon EC2\footnote{https://aws.amazon.com/cn/ec2/instance-types/}. Furthermore, we assume that there exist three types of users: heavy-loaded (type-1), medium-loaded (type-2), and light-loaded (type-3) users. These three types of users may represent  big corporations, medium-size businesses, and small businesses or individual users, respectively. We further assume that the user-type distribution is (10\%, 40\%, 50\%) and the total number of users is 5,000. 

Similar to Section \ref{s4_1_1}, we follow \cite{data_distribution,data_distribution_2}  to generate  bids for each user. We first decide the types of VM in each bid. Specifically, we iterate over all types of VMs and add a type of VM into the bid with probability 80\%. Then, for each type of VM in the bid, we repeatedly add a unit with probability 65\% until a unit is not added or the unit of this VM reaches an upper bound. In Section \ref{s4}, the upper bound is the number of  available units for each item. In cloud computing platforms, the upper bound is generally small to guarantee that the requests of most users can be satisfied. In this paper, we set the upper bound as 5 following \cite{cloud1}. After that, the number of requested units for each type of VMs are multiplied by the corresponding user type factor, denoted as $\rho_i$ for type-$i$ user, to get the actual bundles requested in this bid. Following \cite{cloud1}, we set $\rho_1=2$, $\rho_2=1.5$, and $\rho_3=1$ for the following tests. Finally, the valuation for this bid is randomly picked from 0 and the number of units in this bid.

\subsection{Implementing GNN-Based Method for VM Allocation in Cloud Computing} \label{s5_2}
\begin{figure}
	\centering
	\subfigure[Gap and time.]{
		\begin{minipage}[t]{0.98\linewidth}
			\centering
		\includegraphics[width=0.8\linewidth]{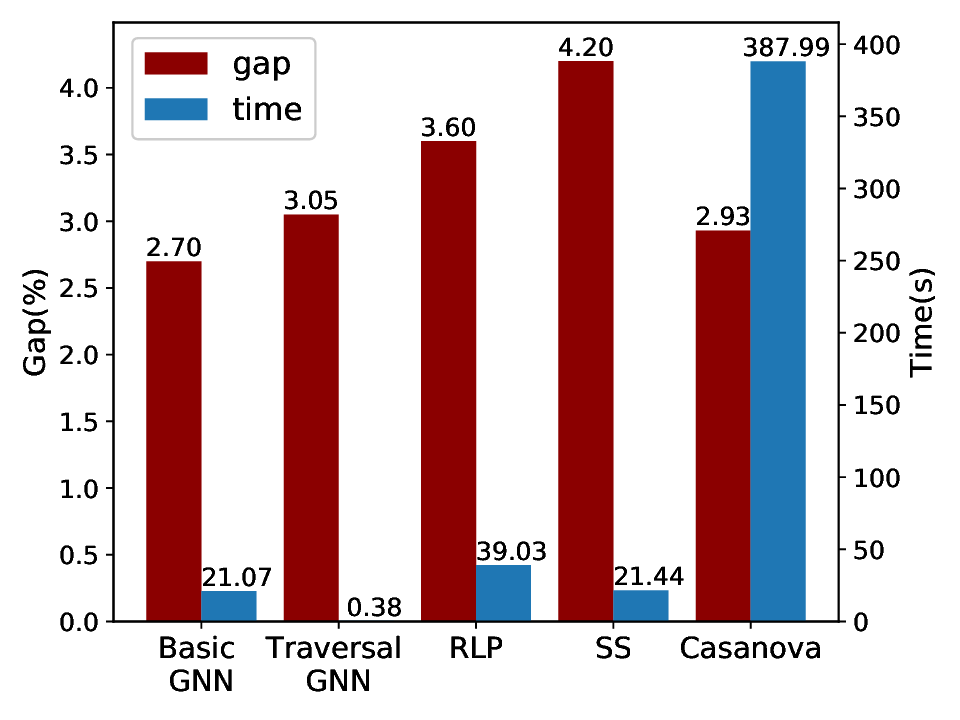}
			\label{cloud_gap_time}
			\vspace{-1em}
		\end{minipage}
	}
	\subfigure[Utilization and satisfaction.]{
		\begin{minipage}[t]{0.98\linewidth}
			\centering
		\includegraphics[width=0.8\linewidth]{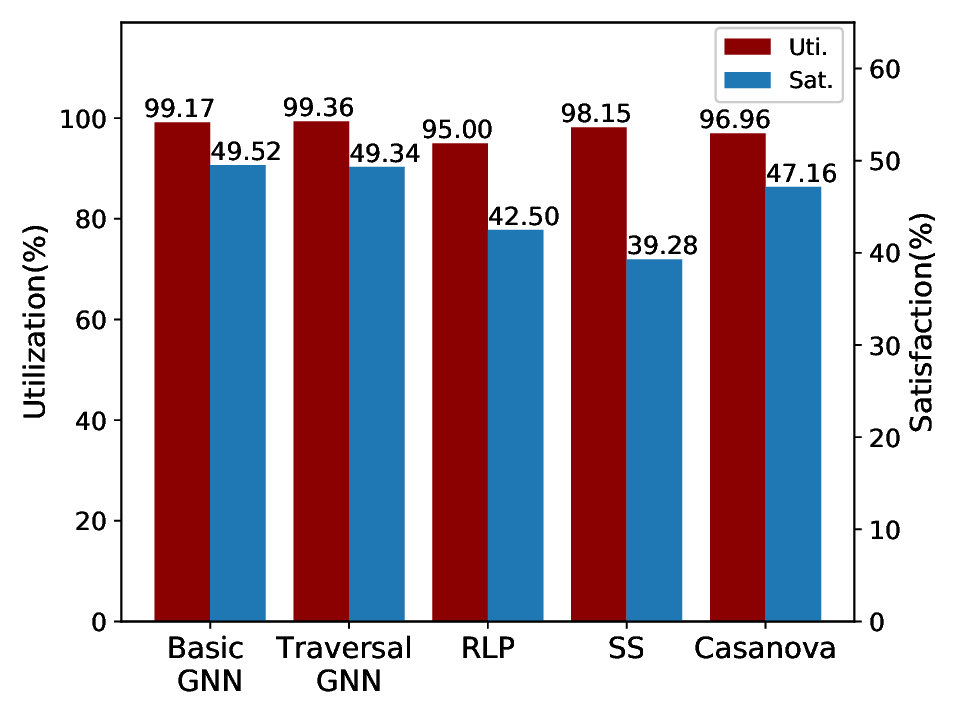}
			\label{cloud_uti_ser}
			\vspace{-1em}
		\end{minipage}
	}
\vspace{-1em}
	\caption{Performance of different methods on VM allocation problem.}
	\label{fig_cloud}
	\vspace{-1em}
\end{figure}

To implement the GNN-based method for the above VM allocation problem with 5,000 users and 90 types of VMs, we first identify the size of the training instances. According to the analysis in Section \ref{s4_6}, the numbers of bids and items of the training instances can be smaller than those of the testing instances. But we need to keep the bid-item ratio, the number of units, and the bid distribution of the training instances consistent with those of the testing instances. Therefore, we use instances with 500 bids, 9 items each with 500 units for training, which can be solved by CPLEX within 1s. Similar to the tests in Section \ref{s4}, we generate 100 random instances for training, 20 for validation, and 60 for testing by the instance generation process mentioned in Section \ref{s5_1}. Then, we set the node keeping probability, $P_k$, as 80\%, and use the optimum-only sample generation process to collect 100,000 samples for training and 20,000 for validation from the training and validation instances, respectively. The testing results are summarized in Table \ref{per_cloud} and Fig. \ref{fig_cloud}.

\begin{table}[htbp]
	\vspace{-1em}
	\scriptsize
	\caption{Performance of Different Methods on VM Allocation Problem}
	\vspace{-1em}
	\label{per_cloud}
	\centering
	\begin{tabular}{|c|c|c|c|c|c|}
		\hline
		Method &\tabincell{c}{Basic \\GNN} &\tabincell{c}{Traversal \\GNN} &RLP  &SS  &Casanova \\
		\hline
		Gap & 2.70\% &3.05\%& 3.60\%&  4.20\%&  2.93\%\\
		\hline
		Time(s)& 21.07&0.38&39.03& 21.44&  387.99\\
		\hline
		Uti.& 99.17\%&99.36\%& 95.00\%&  98.15\%&96.96\%\\
		\hline
		Sat.& 49.52\%&49.34\%&42.50\% &39.28\%&47.16\% \\
		\hline
	\end{tabular}
\end{table}

From Table \ref{per_cloud} and  Fig. \ref{fig_cloud}, our proposed GNN-based method can achieve higher resource utilization and user satisfaction than the benchmarks. Moreover, the basic GNN-based method achieves the smallest performance gap while the traversal GNN-based method has the lowest time complexity. We also notice that performance gaps induced by the RLP and the Casanova are small, which is different from the results in Table \ref{per_heu}. As analyzed in Section \ref{s4_4}, both methods are not suitable for auctions whose available units are limited and user satisfactions are low. Therefore, they perform badly for the synthetic auctions in Section \ref{s4}. On the contrary, in the VM allocation problem,  each type of VMs has many units and the number of required units for each bidder is limited. Therefore, the user satisfaction is high and both the RLP and the Casanova have good performance.

Furthermore, the advantage of the basic GNN-based method in terms of running time is modest when compared with RLP and SS. As mentioned in Section \ref{s3_6_1}, the required iteration number of the basic GNN-based method is positively related to the user satisfaction. For auctions with abundant units, the user satisfactions are generally high, where the basic GNN-based method is not the best choice. Given that the traversal GNN-based method is much faster than the basic one with modest revenue loss (0.35\%), we recommend utilizing the traversal GNN-based method for auctions with abundant available units in practice.

\subsection{Generalization Ability on User-Type Distribution} \label{s5_3}
Generally, the distribution of different types of users will change over the time. For example, type-1 users request a large number of VM units and represent big corporations. They may appear more often during the day time. Therefore, the user-type distributions of the day and night will be different. However, we do not want to change the GNN models frequently with the user-type distributions. Thus, we expect that our proposed method could perform well with the change of user-type distributions over time. In the following, we test the generalization ability of the proposed GNN-based method on user-type distribution. Specifically, we still use the GNN model learned in Section \ref{s5_1} with the user-type distribution  (10\%, 40\%, 50\%). Then we change the distributions of the testing instances and test the performance of the proposed method.  The results are summarized in Table \ref{gene_user_dis}. Comparing Tables \ref{per_cloud} and \ref{gene_user_dis}, the performance gaps and running time are close for instances with different user-type distributions. The results suggest that changing user-type distributions does not influence the performance of the proposed GNN-based method, and thus our proposed method has a good generalization ability in terms of different user-type distributions. 
\begin{table}[htbp]
	\vspace{-1em}
	\scriptsize
	\caption{Generalization Ability of GNN-Based Methods on User-Type Distribution}
	\vspace{-1em}
	\label{gene_user_dis}
	\centering
	\begin{tabular}{|c|c|c|c|}
		\hline
		\diagbox{Distribution}{Gap and Time(s)}{Method} &\tabincell{c}{Basic GNN}&\tabincell{c}{Traversal GNN}\\
		\hline
		(20\%, 30\%, 50\%)&\tabincell{c}{2.69\%\\20.17}&\tabincell{c}{3.12\%\\0.38}\\
		\hline
		(20\%, 40\%, 40\%)& \tabincell{c}{2.94\%\\19.57}&\tabincell{c}{2.92\%\\0.40}\\
		\hline
		(30\%, 30\%, 40\%)&\tabincell{c}{2.85\%\\18.96}& \tabincell{c}{3.13\%\\0.42}\\
		\hline
		(10\%, 10\%, 80\%)&\tabincell{c}{2.86\%\\18.85}& \tabincell{c}{3.14\%\\0.40}\\
		\hline
		(10\%, 30\%, 60\%)&\tabincell{c}{2.85\%\\20.04}& \tabincell{c}{3.13\%\\0.40}\\
		\hline
	\end{tabular}
\vspace{-1.5em}
\end{table}

\section{Conclusions and Future Work} \label{s6}
Multi-unit CA is widely used for resource allocation in different fields, including cloud computing. Different from the traditional auction, the multi-unit CA allows the bidders to submit bids for combinations of items where each item may have more than one unit, and thus enhances the economic efficiency and user flexibility. For a multi-unit CA mechanism, the most difficult and time-consuming part is the multi-unit WDP, which is generally NP-complete to solve and inapproximable. Therefore, this paper incorporates GNNs to efficiently solve the multi-unit WDP with modest revenue loss. By representing the multi-unit WDP as an augmented bipartite  bid-item graph, we  use GNNs to learn a continuous probability map that indicates the probability of each bid belonging to the optimal allocation. To decrease the complexity of the training process, we adopt a special GNN structure with half-convolution operations. Furthermore, given that labeled training instances are difficult to obtain in practice, we propose two different sample generation processes to enhance the sample generation efficiency and decrease the number of needed labeled instances. Finally, two novel graph-based post-processing algorithms are developed to transform the outputs of GNN into feasible solutions of the multi-unit WDP. To verify the effectiveness of the proposed method, we conduct simulations on both  synthetic instances and a specific VM allocation problem in the cloud computing. The results suggest that our proposed method  outperforms CPLEX in terms of time complexity with modest revenue loss. It also outperforms some widely-used  heuristic algorithms in terms of time complexity, revenue loss, resource utilization, and user satisfaction. Meanwhile, it has good generalization ability on some aspects, such as the problem size and the user-type distribution. Therefore, one can start by training a model with small-size instances, and then generalize it to other similar instances with larger sizes and different user-type distributions, which is a preferred property in practice.

The current work represents a first step towards using ML techniques for the multi-unit WDP and some open problems  still need further investigation. First, the two proposed post-processing algorithms need multiple iterations and the inference process of GNN needs to be conducted at each iteration. The running time of the post-processing algorithms is directly influenced by the number of iterations. In this paper, we have proposed the traversal post-processing algorithm to decrease the required iteration number. How to further accelerate the post-processing algorithm is a key point to decrease the complexity of the proposed method. One can also consider leveraging some other  ML techniques, such as reinforcement learning, to efficiently transform the outputs of GNN into feasible solutions. 

Furthermore, our proposed method has limited generalization ability on the number of units, the bid-item ratio, and the bid distribution. To achieve good performance, we need to keep these three parameters of the training and testing instances to be close, which may not be always satisfied in practice. Therefore, enhancing the generalization ability of the proposed method is another important future direction. 

Finally, we note that the GNN-based methodology we have proposed here could potentially be applied to other combinatorial problems in cloud networks. Specifically, the cluster-PM-VM-container allocation problem has been well studied for data centers, with different objectives such as maximizing the energy efficiency and minimizing the execution time. The potential of adapting GNNs to this setting is worth future investigation.

\end{document}